\documentclass[]{article} 
\usepackage{times}
\usepackage{uaiproceed2e}
\usepackage{tikz, pgf}
\usepackage{amsmath, amsthm, bm, amssymb}
\usepackage{algpseudocode, algorithm}
\usepackage{caption,subcaption}
\usepackage[comma, sort&compress]{natbib}
\usepackage{wrapfig}

\DeclareMathOperator*{\argmax}{arg\,max}

\definecolor{DarkRed}{rgb}{0.75,0,0}
\definecolor{DarkGreen}{rgb}{0,0.5,0}
\definecolor{DarkBlue}{rgb}{0,0,0.5}
\definecolor{DarkPurple}{rgb}{0.5,0,0.5}

\allowdisplaybreaks

\title{Approximate Decentralized Bayesian Inference}

\author{
Trevor Campbell \\
LIDS, MIT\\
Cambridge, MA 02139 \\
\texttt{\small tdjc@mit.edu} \\ \rule{.48\textwidth}{0pt}
\And
Jonathan P.\ How \\
LIDS, MIT \\
Cambridge, MA 02139 \\
\texttt{\small jhow@mit.edu} \\\rule{.32\linewidth}{0pt}
}

\begin{document}

\maketitle

\begin{abstract}
This paper presents an approximate method for performing Bayesian inference
in models with conditional independence over a
decentralized network of learning agents. The method first employs variational
inference on each individual learning agent to generate a local approximate posterior, the
agents transmit their local posteriors to other agents in the network, and
finally each agent combines its set of received local posteriors.
The key insight in this work is that, for many Bayesian models,
approximate inference schemes destroy symmetry and dependencies in the model
that are crucial to the correct application of 
Bayes' rule when combining the local posteriors.
The proposed method addresses this issue by including 
an additional optimization step in the combination procedure that
accounts for these broken dependencies. 
Experiments on synthetic and real data demonstrate
that the decentralized method provides advantages in computational performance 
and predictive test likelihood over previous batch and distributed methods.
\end{abstract}
\section{INTRODUCTION}
Recent trends in the growth of datasets, and the methods
by which they are collected, have led to increasing
interest in the parallelization of machine learning
algorithms. Parallelization results in reductions in both the memory usage and computation
time of learning, and allows data to be collected by a network of learning
agents rather than by a single central agent.
There are two major classes of parallelization algorithms: those that require a
globally shared memory/computation unit (e.g., a central fusion
processor that each learning agent is in communication with, or the main
thread on a multi-threaded computer), and those that do not.
While there is as of yet no consensus in the literature on
the terminology for these two types of parallelization, in this work we refer to
these two classes, respectively, as 
\emph{distributed} and \emph{decentralized} learning.

Some recent approaches to distributed learning have involved using streaming variational
approximations~\citep{Broderick13_NIPS}, parallel stochastic gradient
descent \citep{Niu11_NIPS}, the Map-Reduce framework \citep{Dean04_OSDI}, 
database-inspired concurrency control \citep{Pan13_NIPS},
 and message passing on
graphical models~\citep{Gonzalez09_AISTATS}. When a reliable central
learning agent with sufficient communication bandwidth is available, such distributed learning
techniques are generally preferred to decentralized learning. This is a result
of the consistent global model shared by all agents, with which they can
make local updates without the concern of generating conflicts unbeknownst to
each other.

Decentralized learning is a harder problem in general, due to asynchronous
communication/computation, a lack of a globally shared state, and potential network
and learning agent failure, all of which may lead to inconsistencies in
the model possessed by each agent. Addressing these issues is particularly relevant to
mobile sensor networks in which the network structure
varies over time, agents drop out and are added dynamically, and no single
agent has the computational or communication resources to act as a central hub
during learning. Past approaches to decentralized
learning typically involve each agent communicating frequently to form a consensus on the model over the
network, and are often model-specific: particle 
filtering for state estimation~\citep{Rosencrantz03_UAI} involves
sending particle sets and informative measurements to peers; distributed EM~\citep{Wolfe08_ICML} 
requires communication of model statistics to the network after each local iteration; distributed Gibbs
sampling~\citep{Newman07_NIPS} involves model synchronization after each sampling
step; robust distributed inference~\citep{Paskin04_UAI} requires the formation
of a spanning tree of nodes in the network and message passing;
asynchronous distributed learning of topic models~\citep{Asuncion08_NIPS} requires communication of 
model statistics to peers after each local sampling step; and hyperparameter
consensus~\citep{Fraser12_AUTO} requires using linear network consensus on
exponential family hyperparameters. 

The method proposed in the present paper takes a different tack; each agent
computes an approximate factorized variational posterior using only their local
datasets, sends and receives statistics to and from other agents in the network
asynchronously, and combines the posteriors locally on-demand. Building upon insights from previous work on distributed and
decentralized inference~\citep{Broderick13_NIPS,Rosencrantz03_UAI}, a na\"{i}ve
version of this algorithm based on Bayes' rule is presented. It is then
shown that, due to the approximation used in variational inference,
this algorithm leads to poor decentralized
posterior approximations for unsupervised models with inherent symmetry.
Next, building on insights gained from the results of variational and Gibbs
sampling inference on a synthetic example,
an approximate posterior combination algorithm is presented
that accounts for symmetry structure in models that the
na\"{i}ve algorithm is unable to capture. The proposed method is highly
flexible, as it can be combined with past streaming variational
approximations~\citep{Lin13_NIPS,Broderick13_NIPS}, agents can 
share information with only subsets of the network, the network may be dynamic
with unknown toplogy, and the failure of
individual learning agents does not affect the operation of the rest of the
network. Experiments on a mixture model, latent Dirichlet
allocation~\citep{Blei03_JMLR}, and latent feature
assignment~\citep{Griffiths05_NIPS}
demonstrate that the decentralized method provides advantages in model performance
and computational time over previous approaches.

\section{APPROXIMATE DECENTRALIZED BAYESIAN INFERENCE}
\subsection{THE NA\"{I}VE APPROACH}
Suppose there is a set of learning agents $i$, $i=1, \dots, N$, each with a distribution on a set 
of latent parameters $\theta_j$, $j=1, \dots, K$ (all parameters $\theta_{j}$ may generally be vectors). 
Suppose a fully factorized exponential family distribution has been
used to approximate each agent's posterior $q_i(\theta_1, \dots, \theta_K)$. Then the
distribution possessed by each agent $i$ is
\begin{align}
  q_i(\theta_{1}, \dots, \theta_{K}) &= \prod_j
  q_{\lambda_{ij}}(\theta_{j}),
\end{align}
where $\lambda_{ij}$ parameterizes agent $i$'s distribution over $\theta_j$.
Given the prior
\begin{align}
  q_0(\theta_1, \dots, \theta_K) &= \prod_j q_{\lambda_{0j}}(\theta_j),
\end{align}
is known by all agents, and the conditional independence of data given the
model, the overall posterior distribution $q(\theta_1, \dots, \theta_K)$ 
 may be approximated
by using Bayes' rule~\citep{Broderick13_NIPS} and summing over the $\lambda_{ij}$:
\begin{align}
  \begin{aligned}
  q(\cdot) &\propto q_0(\cdot)^{1-N}\prod_i q_i(\cdot)\\
  &= \left(\prod_j q_{\lambda_{0j}}(\theta_j)\right)^{1-N} \prod_i \prod_j
  q_{\lambda_{ij}}(\theta_j)\\
  \therefore q(\cdot) &= \prod_j q_{\lambda_j}(\theta_j)\\
 \text{where }\lambda_j &= (1-N)\lambda_{0j}+\sum_i \lambda_{ij}.
  \end{aligned}\label{eq:decbayes}
\end{align}
The last line follows from the use of exponential family distributions in
the variational approximation. This procedure is decentralized, as each agent
can asynchronously compute its individual posterior approximation, broadcast it to the network, 
receive approximations from other agents, and combine them locally. 
 Furthermore,
this procedure can be made to handle streaming data by using a technique such as SDA-Bayes~\citep{Broderick13_NIPS} or
sequential variational approximation~\citep{Lin13_NIPS} on
each agent locally to generate the streaming local posteriors $q_i$. 

\begin{figure*}[t!]
  \captionsetup{font=scriptsize}
  \centering
  \begin{subfigure}[b]{0.4\textwidth}
    \includegraphics[width=\columnwidth]{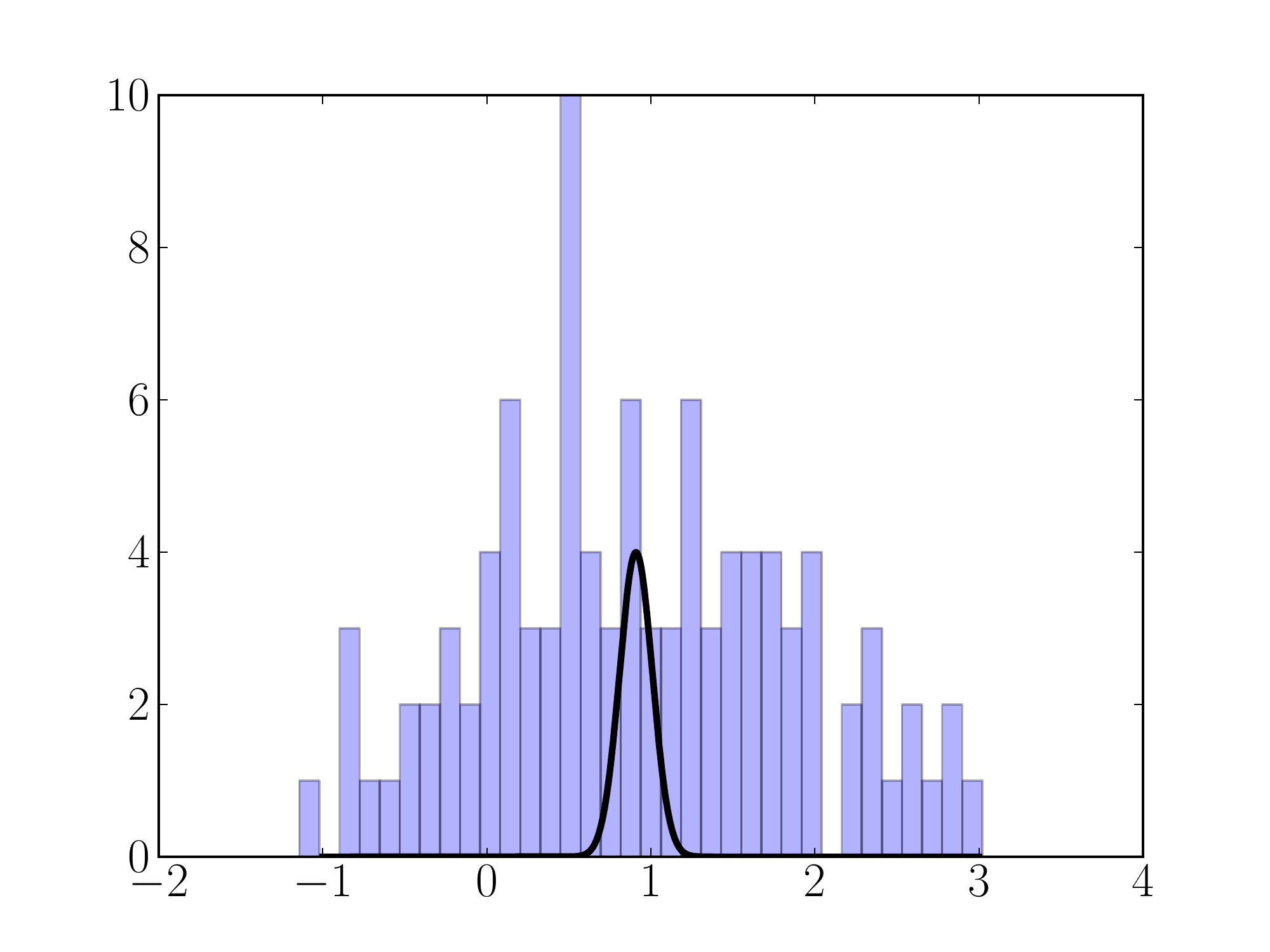}\vspace*{-.15in}
    \caption{}\label{fig:postgausscent}
  \end{subfigure}
  \begin{subfigure}[b]{0.4\textwidth}
    \includegraphics[width=\columnwidth]{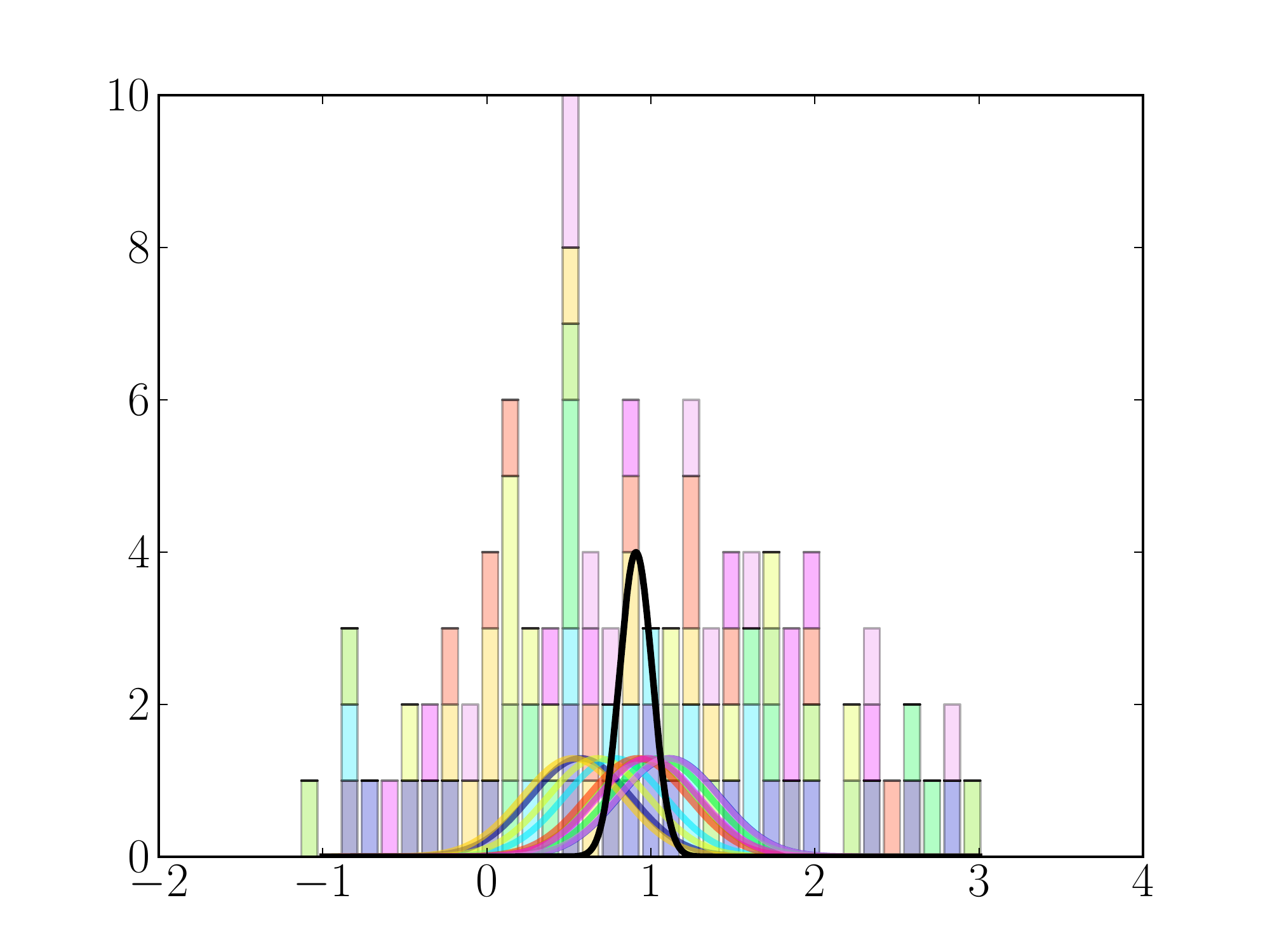}\vspace*{-.15in}
    \caption{}\label{fig:postgaussdec}
  \end{subfigure}
\vspace*{-.1in}
  \caption{(\ref{fig:postgausscent}): Batch posterior of $\mu$ in black,
  with histogram of observed data.
  (\ref{fig:postgaussdec}): Decentralized posterior of
$\mu$ in black, individual posteriors in color and correspondingly colored histogram
of observed data.}\label{fig:postgauss}
\end{figure*}

As an example, this method is now applied to decentralized learning of 
a Gaussian model with
unknown mean $\mu = 1.0$ and known variance $\sigma^2 = 1.0$. The prior on $\mu$
is Gaussian with mean $\mu_0 = 0.0$ and variance $\sigma^2_0 = 2.0$. There are
10 learning agents, each of whom receives 10 observations $y \sim
\mathcal{N}(\mu, \sigma^2)$. Because the Gaussian distribution is in the
exponential family, the variational approximation is exact in this case.
As shown in Figure \ref{fig:postgauss}, the decentralized posterior is
the same as the batch posterior. Note that if the approximation
is used on a more complicated distribution not in the exponential family,
then the batch posterior may in general differ from the decentralized
posterior; however, they will both approximate the same true posterior
distribution.

\subsection{FAILURE OF THE NA\"{I}VE APPROACH UNDER PARAMETER PERMUTATION SYMMETRY}
\begin{figure*}[t!]
  \captionsetup{font=scriptsize}
  \centering
  \begin{subfigure}[b]{0.4\textwidth}
    \includegraphics[width=\columnwidth]{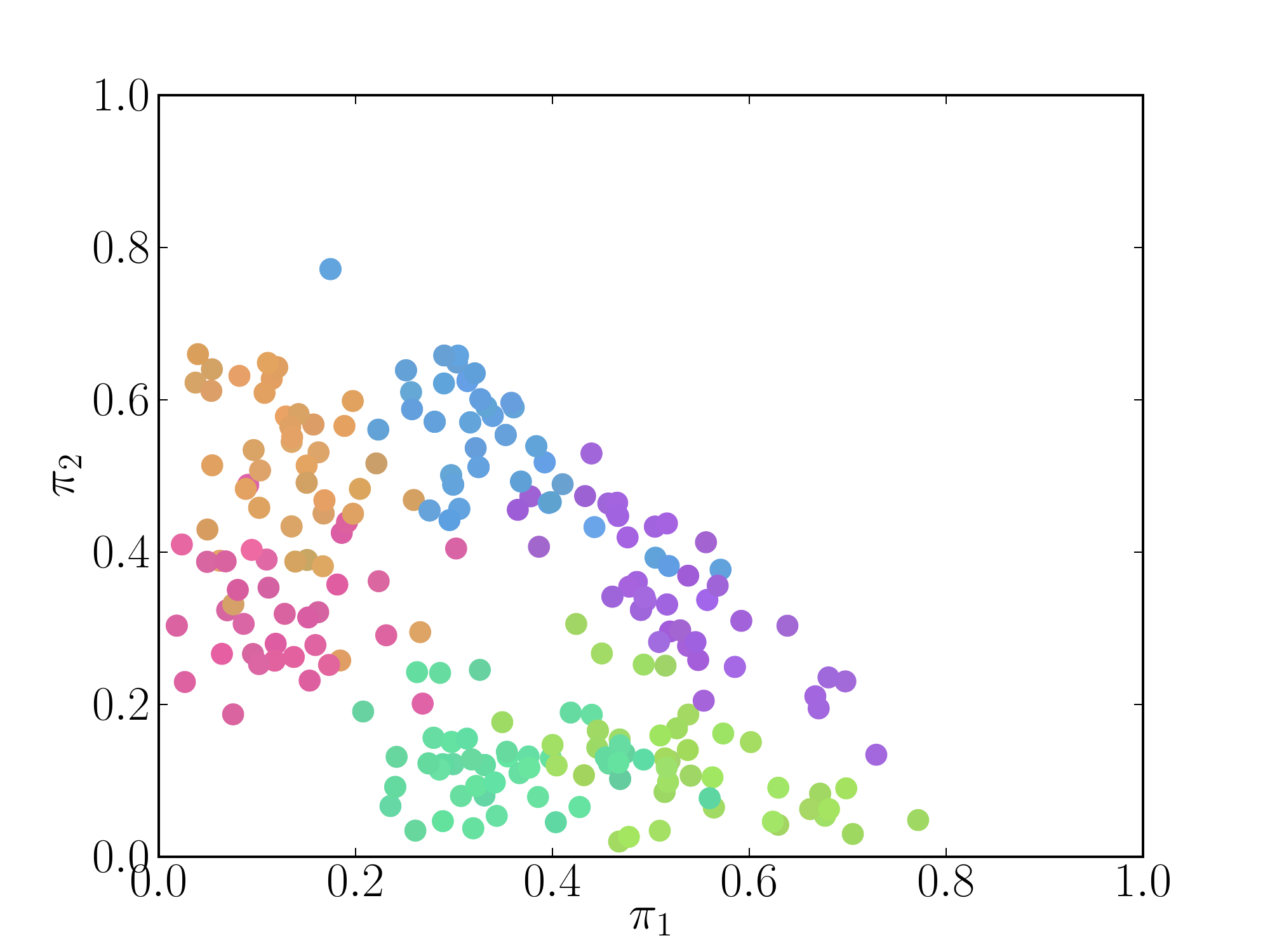}\vspace*{-.1in}
    \caption{}\label{fig:postmixtrue}
  \end{subfigure}
  \begin{subfigure}[b]{0.4\textwidth}
    \includegraphics[width=\columnwidth]{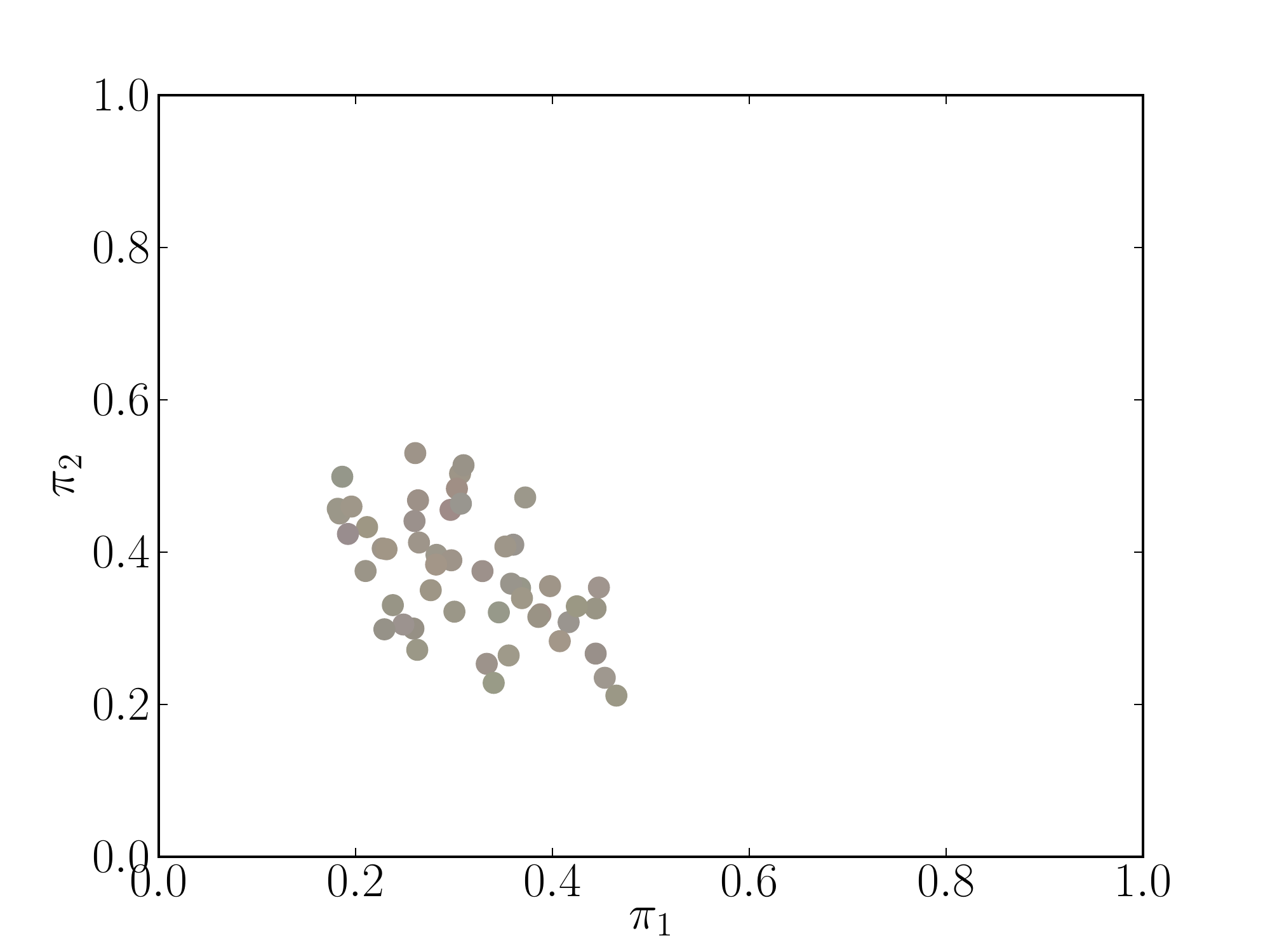}\vspace*{-.1in}
    \caption{}\label{fig:postmixdecbad}
  \end{subfigure}
  \caption{(\ref{fig:postmixtrue}): Samples from the true posterior over $\mu, \pi$.
  Each particle's position on the simplex (with $\pi_3 = 1-\pi_1-\pi_2$)
  represents the sampled weights, while RGB color coordinates of each particle
  represent the sampled position of the three means.
  (\ref{fig:postmixdecbad}): Samples from the na\"{i}vely constructed decentralized approximate
posterior, with the same coloring scheme. Note the disparity with Figure
\ref{fig:postmixtrue}.}\label{fig:postmix}
\end{figure*}

As a second example, we apply decentralized inference to a Gaussian mixture
model with three components having unknown means $\mu = (1.0, -1.0, 3.0)$
and cluster weights $\pi = (0.6, 0.3, 0.1)$ with known variance $\sigma^2 =
0.09$. The prior on each mean $\mu_i$ was Gaussian with mean $\mu_0 = 0.0$ and
variance $\sigma^2_0 = 2.0$, while the prior on the weights $\pi$ 
was Dirichlet with parameters $(1.0, 1.0, 1.0)$. 
First, the true posterior, shown in Figure \ref{fig:postmixtrue}, was formed using 30 datapoints
that were sampled from the generative model. Then, the decentralized variational
inference procedure in (\ref{eq:decbayes}) was
run with 10 learning agents, each of whom received 3 of the datapoints,
resulting in the approximate decentralized posterior in Figure \ref{fig:postmixdecbad}.

The decentralized posterior, in this case, is a very poor approximation
of the true batch posterior.  The reason for this is straightforward:
approximate inference algorithms, such as variational inference with a
fully factorized distribution,
often do not capture \emph{parameter
permutation symmetry} in the posterior. Parameter permutation symmetry
is a property of a Bayesian model in which permuting the values of 
some subset of the parameters does not change the posterior probability.
For example, in the Gaussian mixture model, the true posterior
over $\pi, \mu$ given data $y$ is invariant to transformation by any permutation matrix $P$:
\begin{align}
  p(P\pi, P\mu | y) &= p(\pi, \mu | y).
\end{align}
Indeed, examining the true posterior in Figure \ref{fig:postmixtrue}, one can identify 6
differently colored regions; each of these regions corresponds to one of the
possible $3! = 6$ permutation matrices $P$. In other words, the true posterior
captures the invariance of the distribution to reordering of the parameters correctly.

\begin{figure*}[t!]
\vspace*{-.1in}
  \captionsetup{font=scriptsize}
  \centering
  \begin{subfigure}[t]{0.19\textwidth}
    \includegraphics[width=\columnwidth]{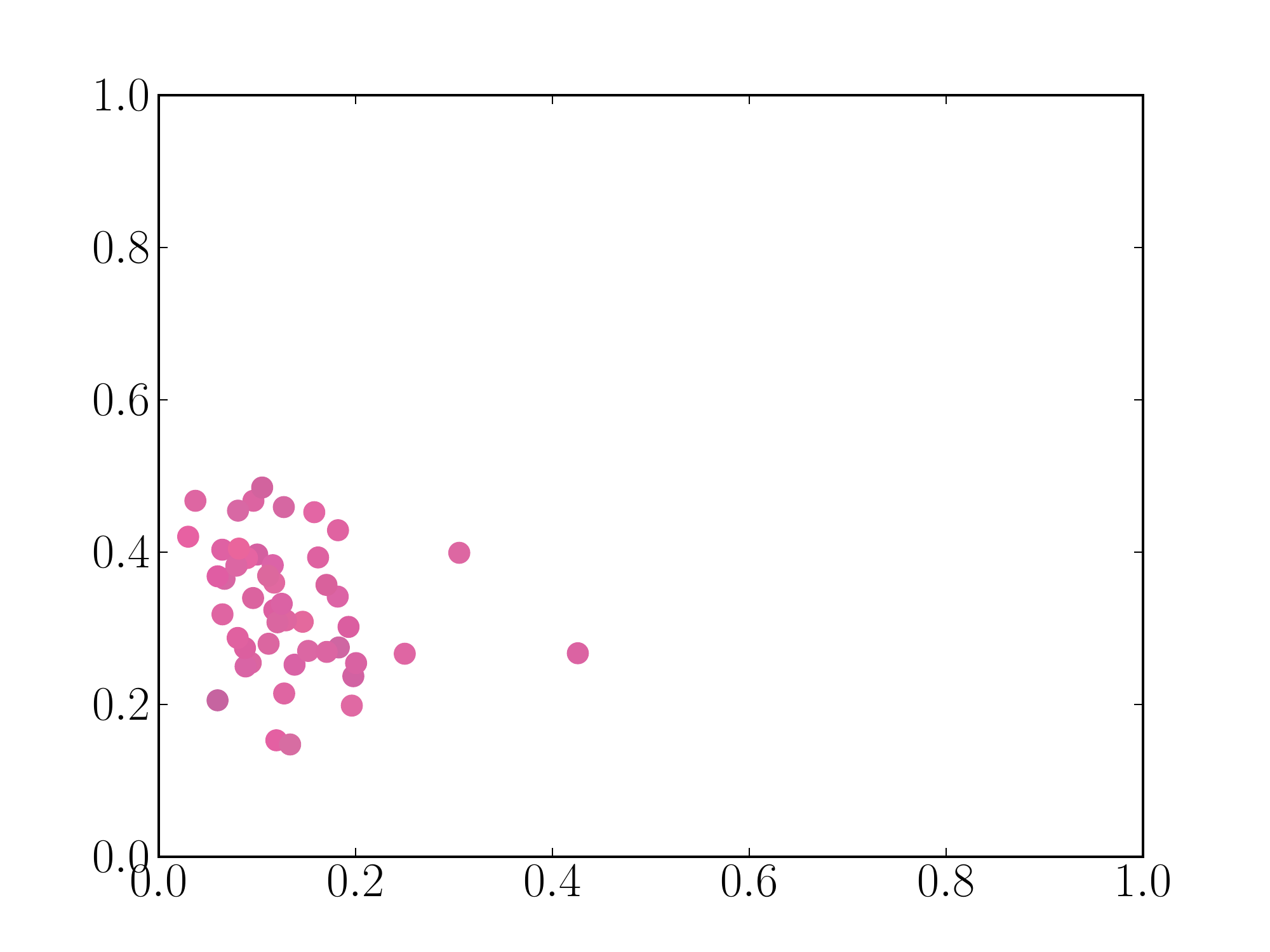}\vspace*{-.1in}
  \end{subfigure}
  \begin{subfigure}[t]{0.19\textwidth}
    \includegraphics[width=\columnwidth]{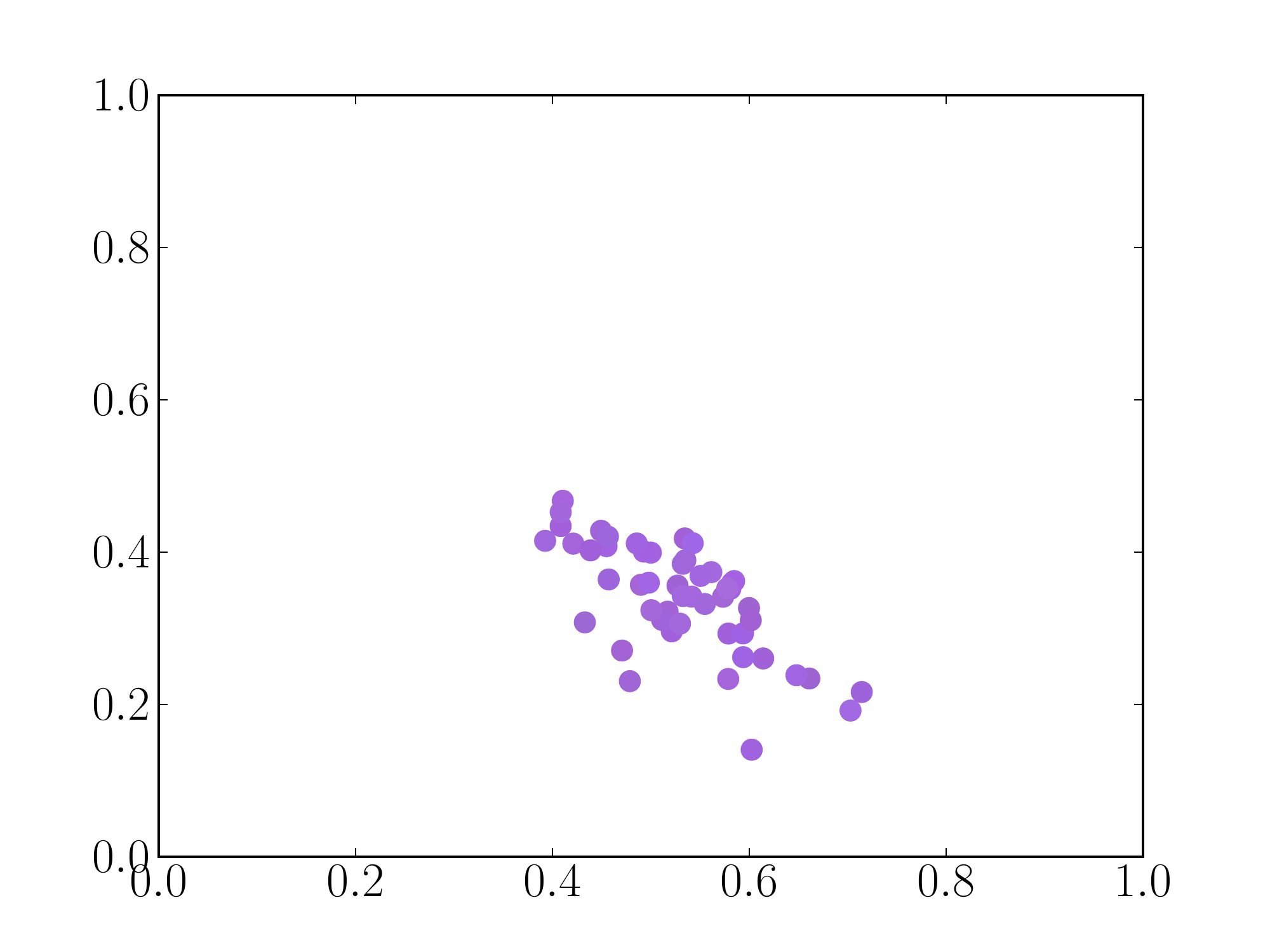}\vspace*{-.1in}
  \end{subfigure}
  \begin{subfigure}[t]{0.19\textwidth}
    \includegraphics[width=\columnwidth]{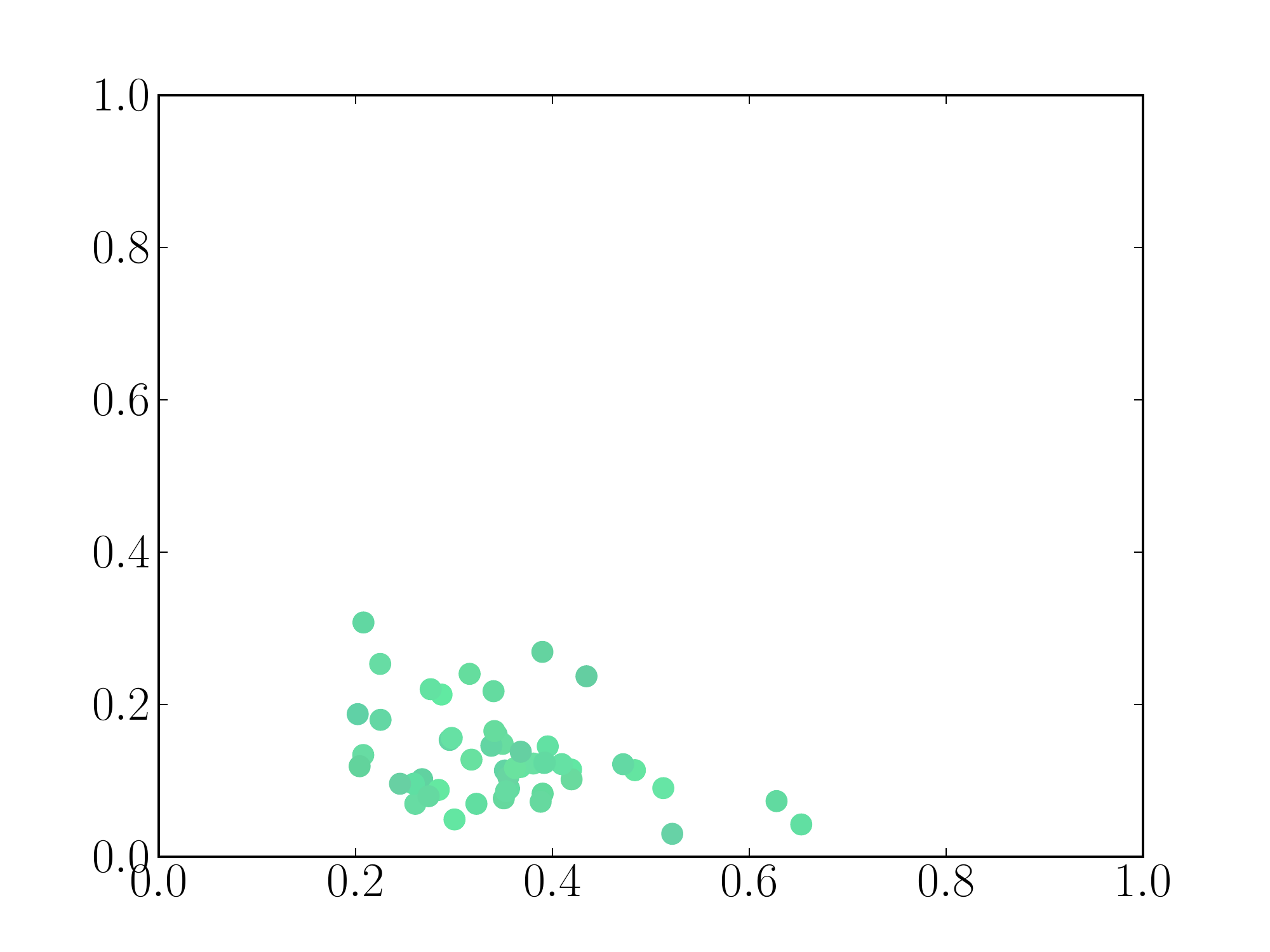}\vspace*{-.1in}
    \caption{}\label{fig:postmixgibbs}
  \end{subfigure}
  \begin{subfigure}[t]{0.19\textwidth}
    \includegraphics[width=\columnwidth]{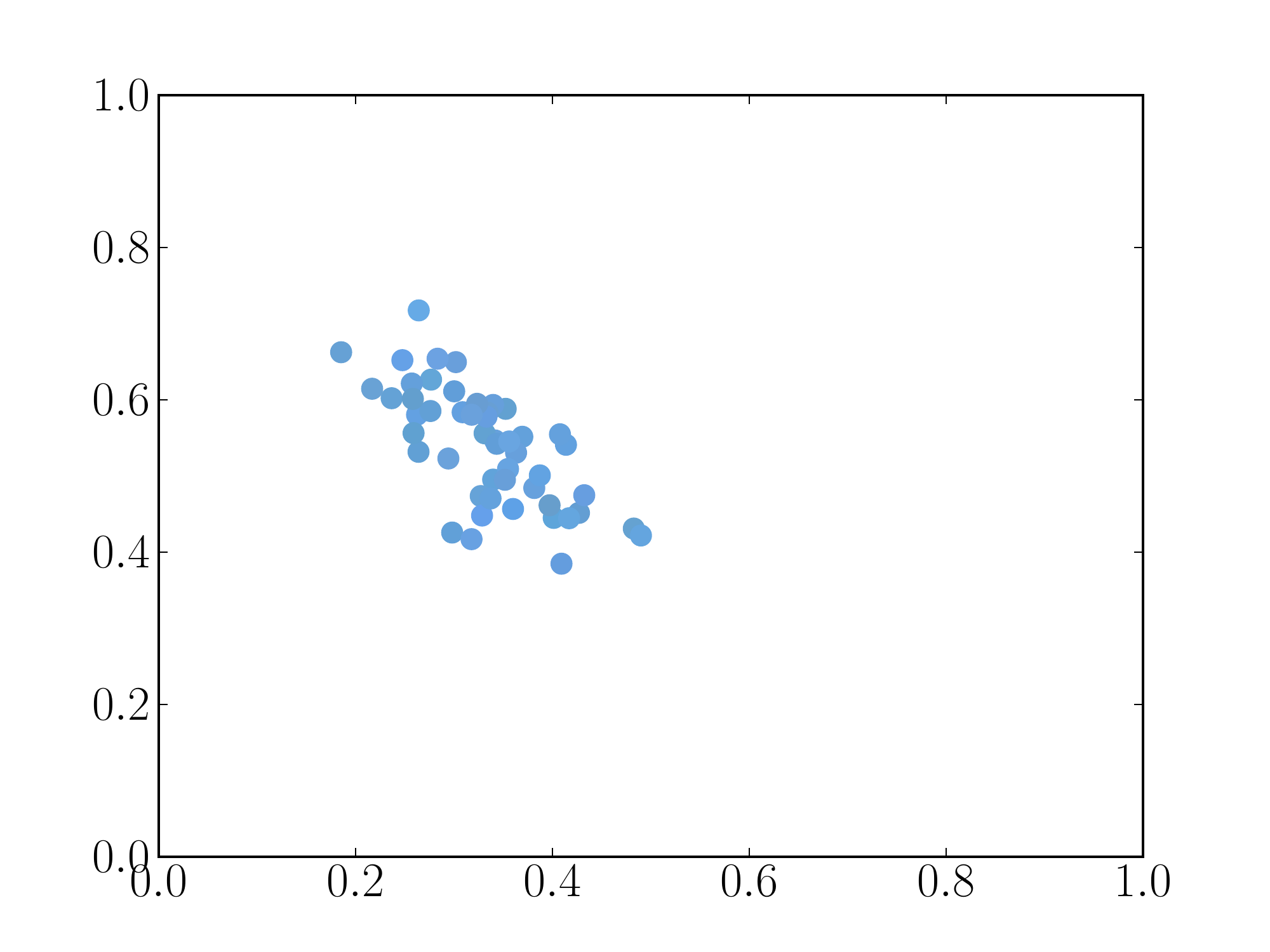}\vspace*{-.1in}
  \end{subfigure}
  \begin{subfigure}[t]{0.19\textwidth}
    \includegraphics[width=\columnwidth]{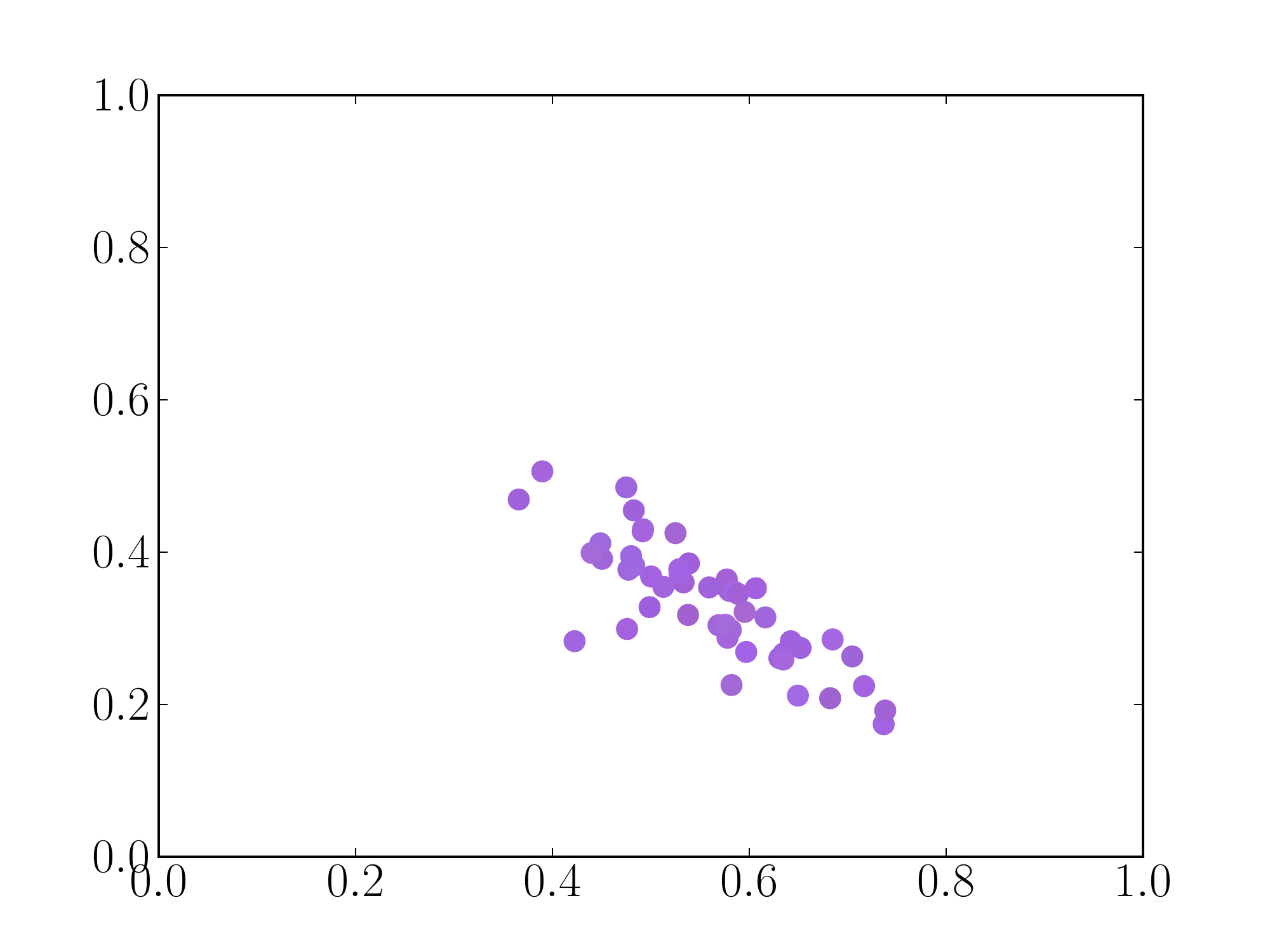}\vspace*{-.1in}
  \end{subfigure}
  %
  \begin{subfigure}[t]{0.19\textwidth}
    \includegraphics[width=\columnwidth]{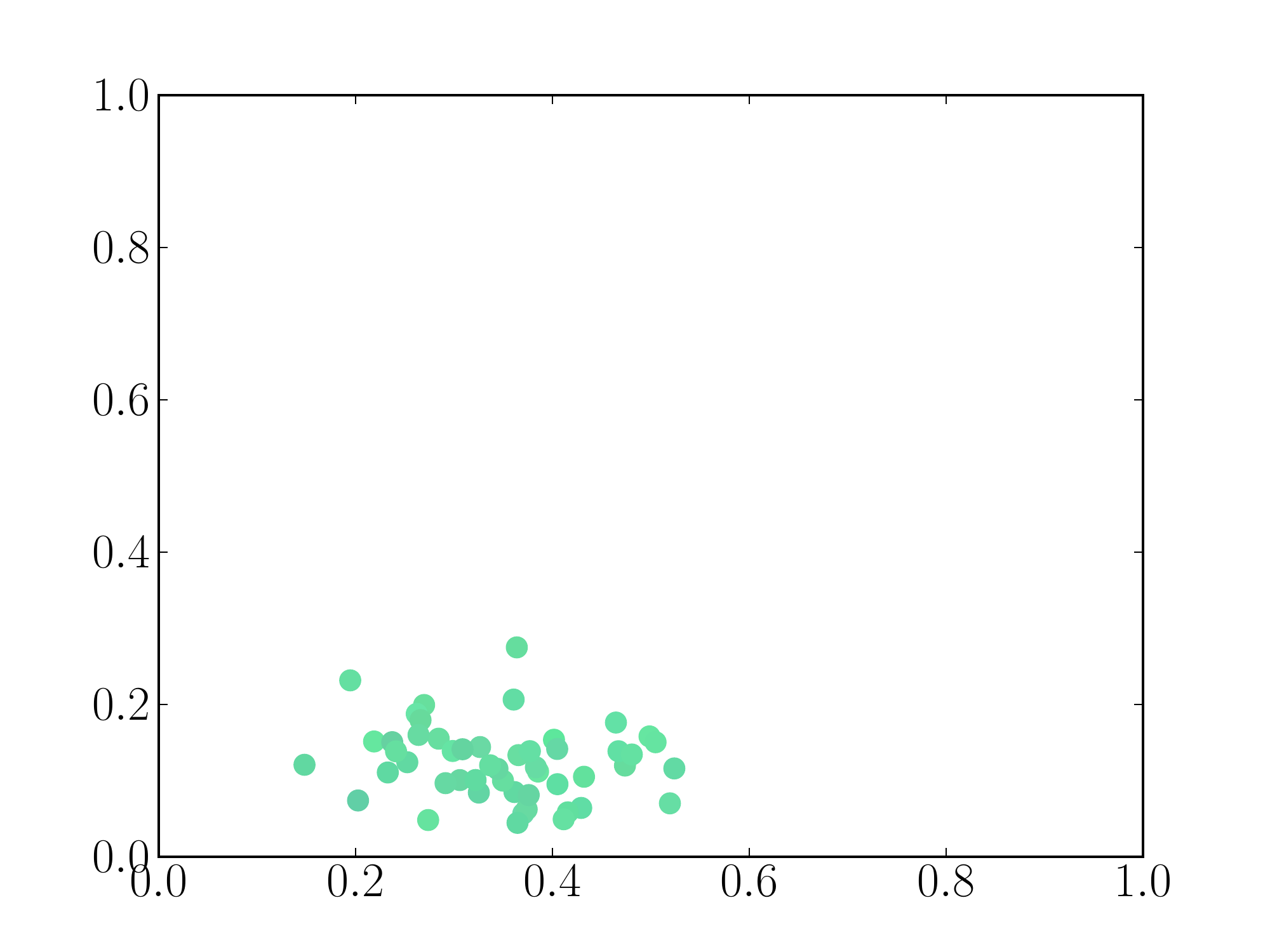}\vspace*{-.1in}
  \end{subfigure}
  \begin{subfigure}[t]{0.19\textwidth}
    \includegraphics[width=\columnwidth]{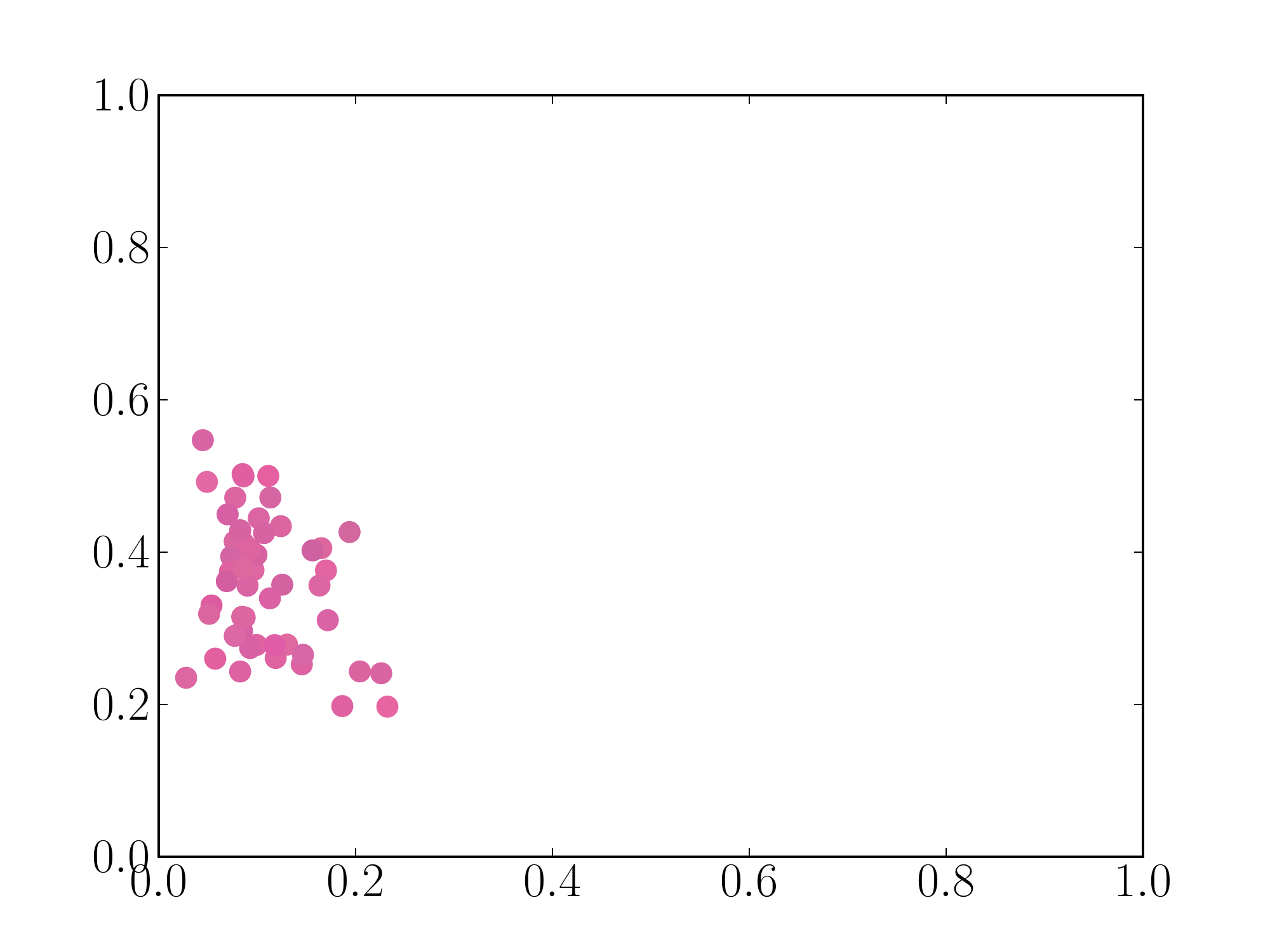}\vspace*{-.1in}
  \end{subfigure}
  \begin{subfigure}[t]{0.19\textwidth}
    \includegraphics[width=\columnwidth]{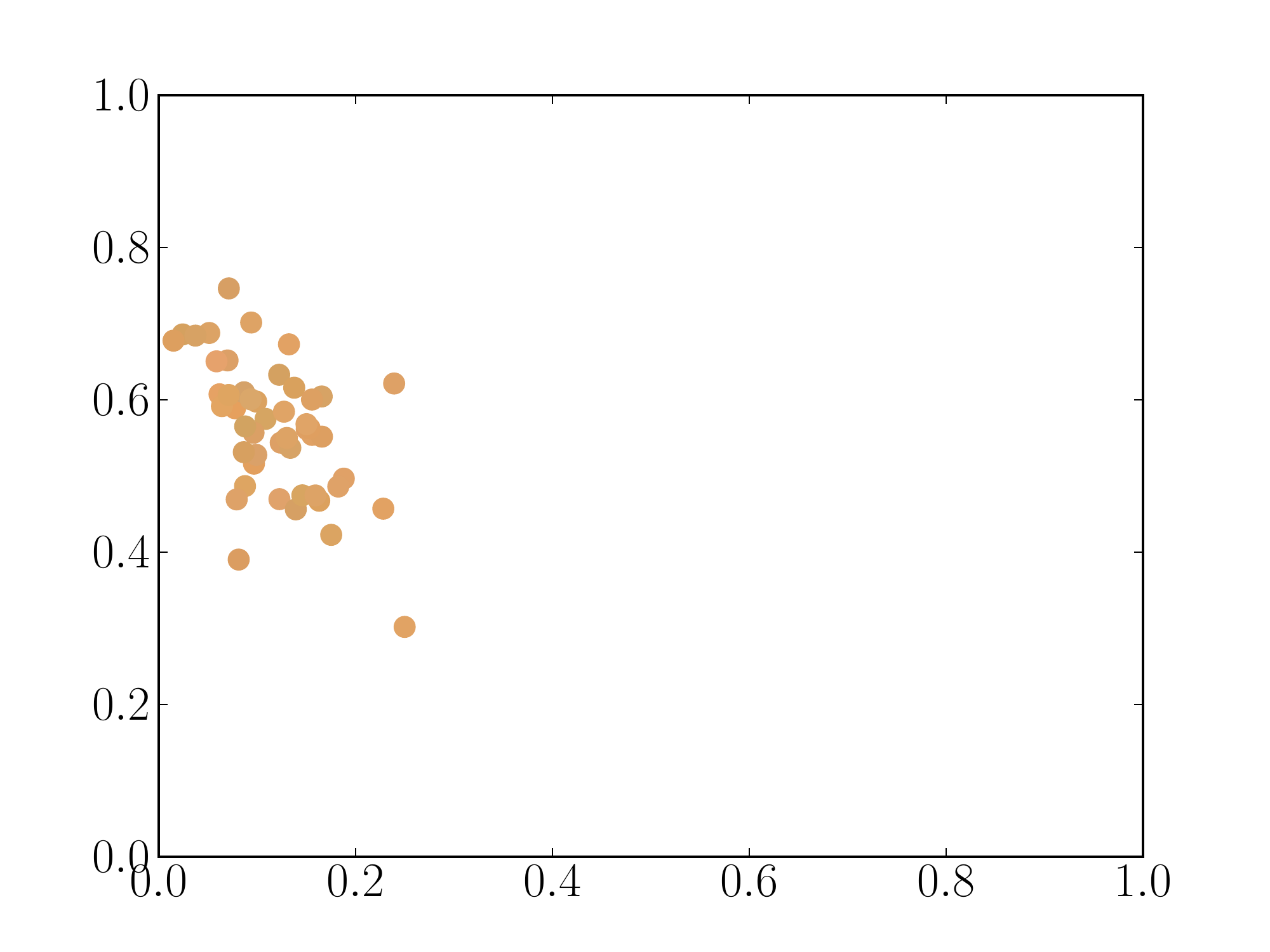}\vspace*{-.1in}
    \caption{}\label{fig:postmixvb}
  \end{subfigure}
  \begin{subfigure}[t]{0.19\textwidth}
    \includegraphics[width=\columnwidth]{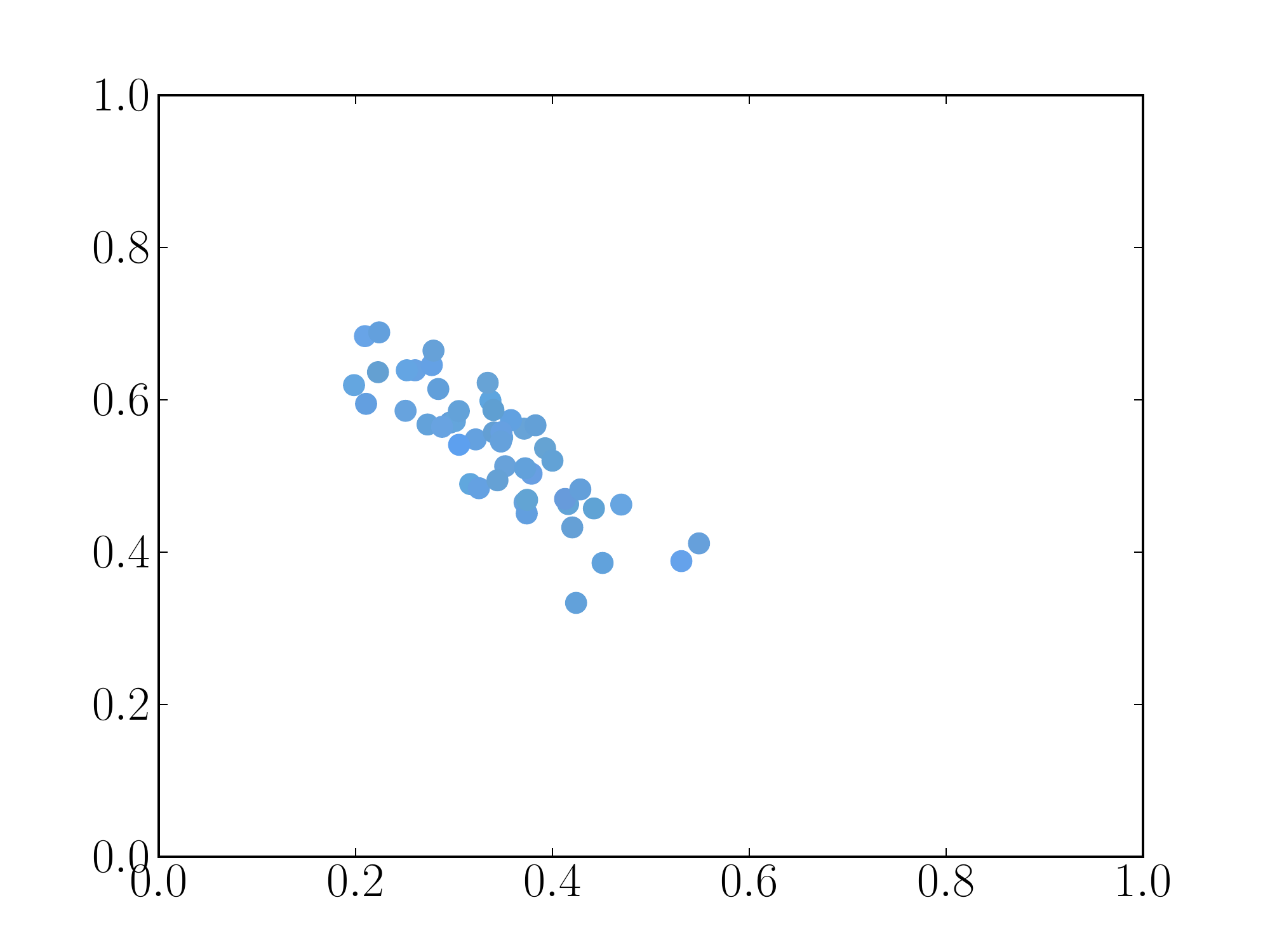}\vspace*{-.1in}
  \end{subfigure}
  \begin{subfigure}[t]{0.19\textwidth}
    \includegraphics[width=\columnwidth]{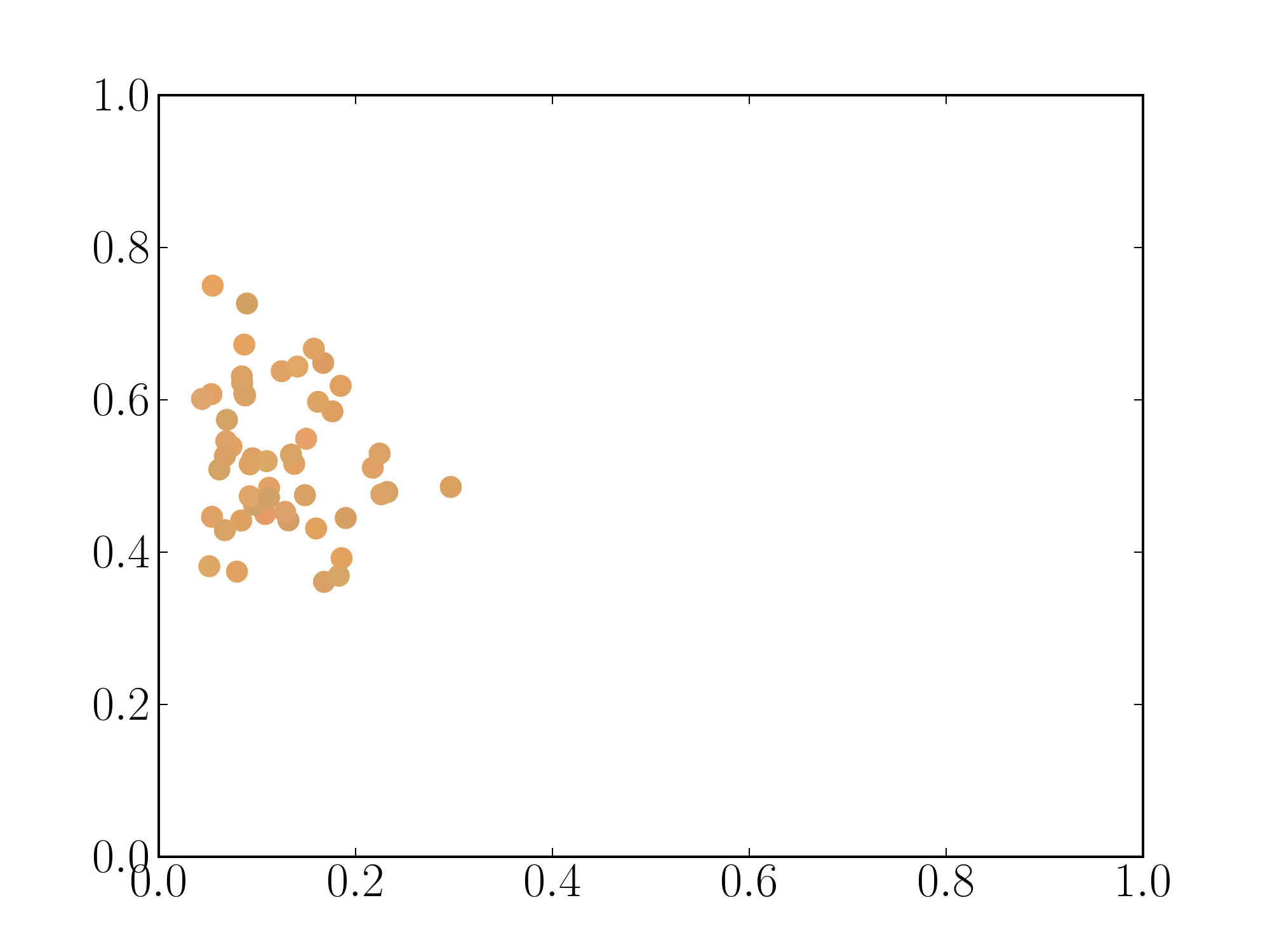}\vspace*{-.1in}
  \end{subfigure}
  %
\vspace*{-.1in}
  \caption{Batch Gibbs sampling (\ref{fig:postmixgibbs})  and variational Bayes
  (\ref{fig:postmixvb}) 
  approximate posterior samples from 5 random restarts. Comparison to Figure
  \ref{fig:postmixtrue} shows that both approximate inference algorithms tend
  to converge to a random component of the permutation symmetry in the true
  posterior.
}\label{fig:postmixapprox}
\end{figure*}

To demonstrate that approximate inference algorithms typically do not capture
parameter permutation symmetry in a model, consider 
the same mixture model, learned with 30 datapoints in a single batch
using Gibbs sampling and variational Bayesian inference. Samples from
5 random restarts of each method are shown in Figure \ref{fig:postmixapprox}.
Both algorithms fail to capture the permutation symmetry in the mixture
model, and converge to one of the 6 possible orderings of the parameters.
This occurs in variational Bayesian inference and Gibbs sampling for different
reasons: Gibbs sampling algorithms often get stuck in local posterior likelihood
optima, while the variational approximation explicitly breaks 
the dependence of the parameters on one another. 

In a batch setting,
this does not pose a problem, because practitioners typically find the selection
of a particular parameter ordering acceptable.  However, in the decentralized
setting, this causes problems when combining the posteriors of individual
learning agents. If each agent effectively picks a parameter ordering at random when
performing inference, combining the posteriors without considering those
orderings can lead to poor results (such as that presented in Figure
\ref{fig:postmixdecbad}). Past work dealing with this issue has focused
primarily on modifying the samples of MCMC algorithms by
introducing ``identifiability constraints'' that control the ordering
of parameters~\citep{Stephens00_JRSSB,Jasra05_SS}, but these approaches are
generally model-specific and restricted to use on very simple mixture models.

\subsection{MERGING POSTERIORS WITH PARAMETER PERMUTATION SYMMETRY}
This section presents a method for locally combining the individual posteriors of
decentralized learning agents when the model contains parameter
permutation symmetry. Formally, suppose that
the true posterior probability of $\theta_1, \dots, \theta_K$ is invariant to permutations
of the components of one or more $\theta_j$. In general, there may be subsets of
parameters which have coupled symmetry,
 in that the true posterior is only invariant
if the components of all parameters in the subset are permuted in the same way (for example,
the earlier Dirichlet mixture model had coupled permutation symmetry in $\mu$
and $\pi$). It is assumed that any such coupling in the model is known beforehand by all
agents. Because the exponential family variational approximation is
completely decoupled, it is possible to treat each coupled permutation symmetry
set of parameters in the model independently; therefore, we assume below that
$\theta_1, \dots, \theta_K$ all have coupled permutation symmetry, for
simplicity in the notation and exposition.

\begin{figure*}
  \captionsetup{font=scriptsize}
  \centering
  \begin{subfigure}[t]{0.19\textwidth}
    \includegraphics[width=\columnwidth]{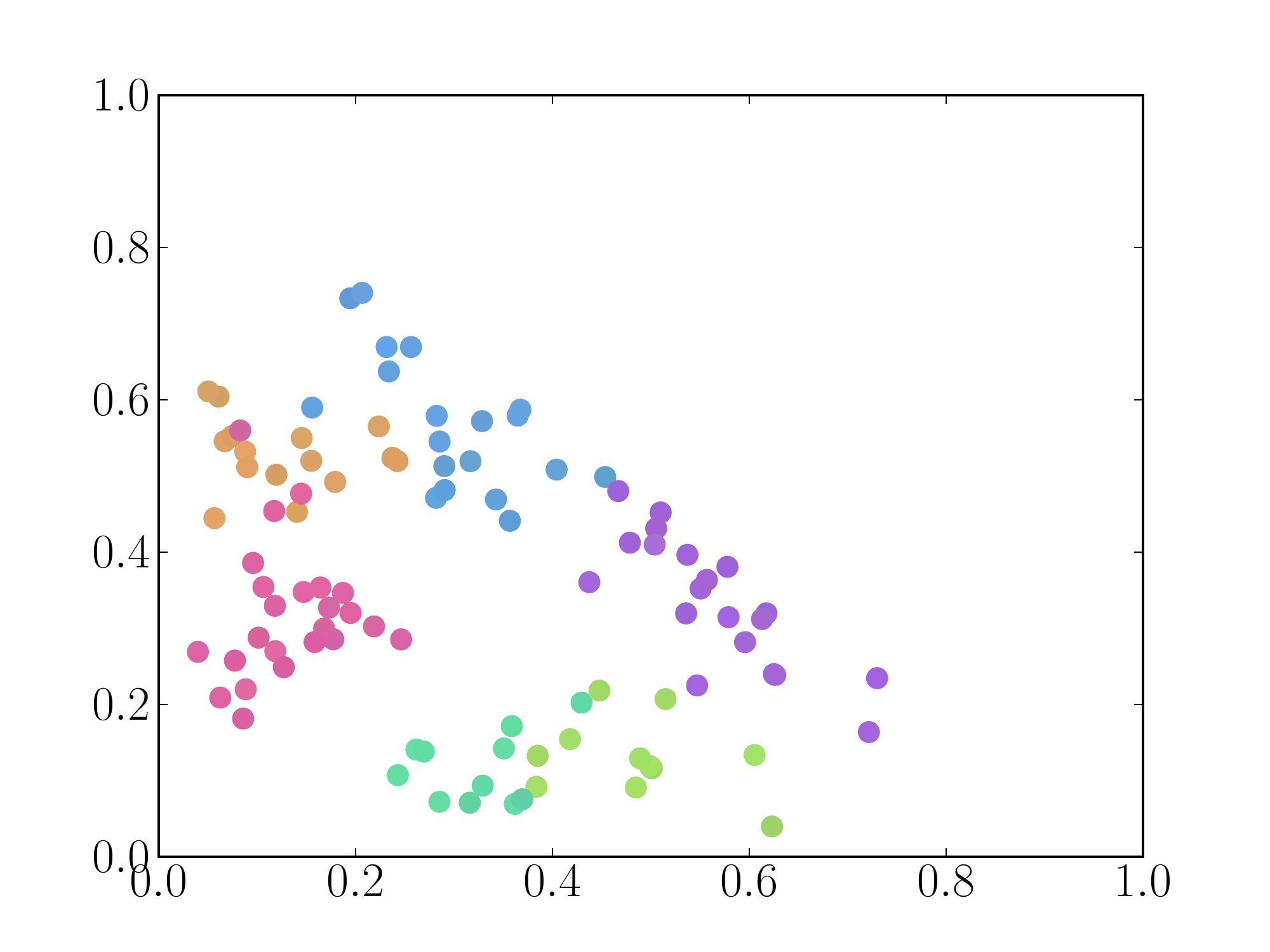}
  \end{subfigure}
  \begin{subfigure}[t]{0.19\textwidth}
    \includegraphics[width=\columnwidth]{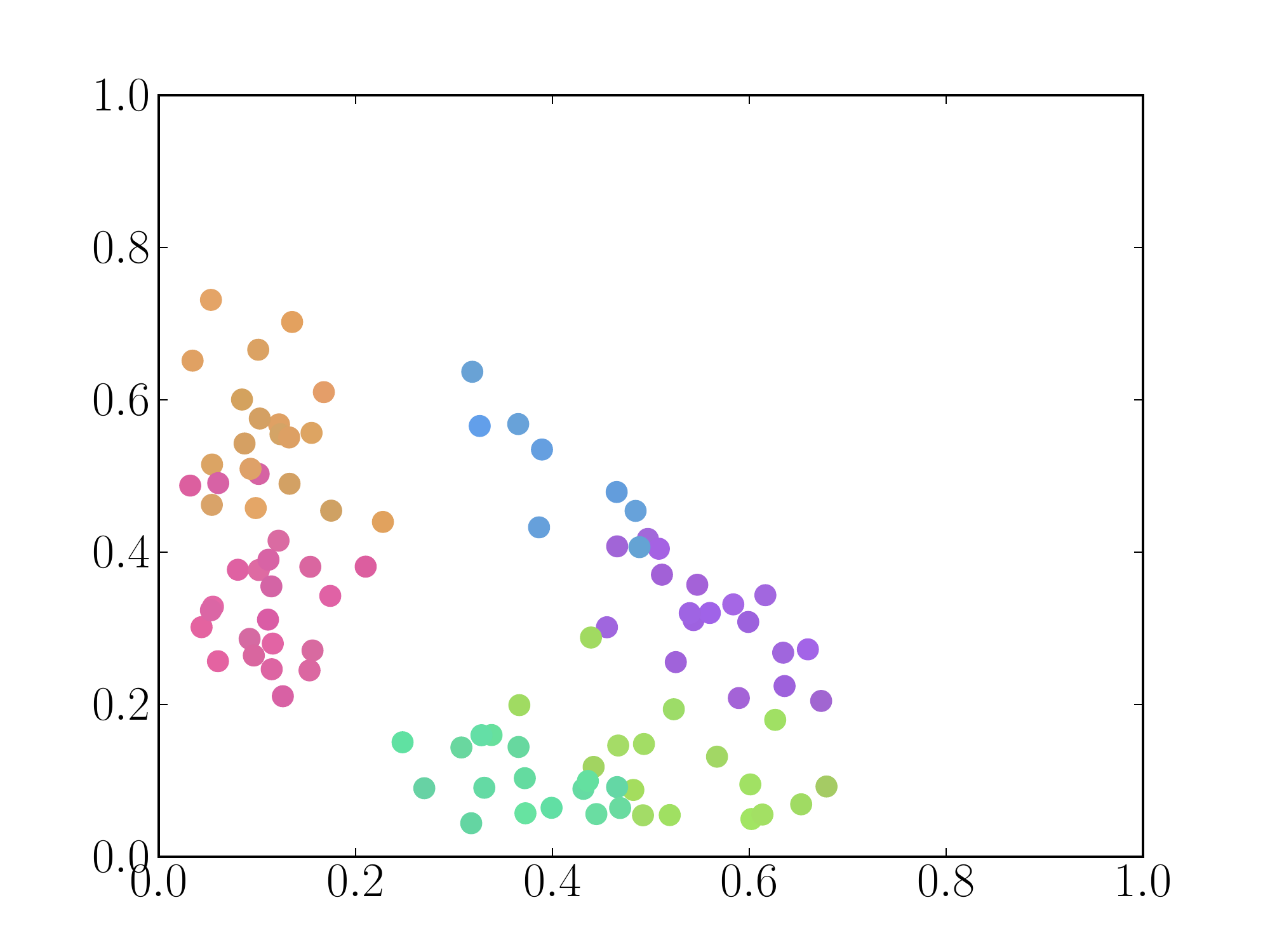}
  \end{subfigure}
  \begin{subfigure}[t]{0.19\textwidth}
    \includegraphics[width=\columnwidth]{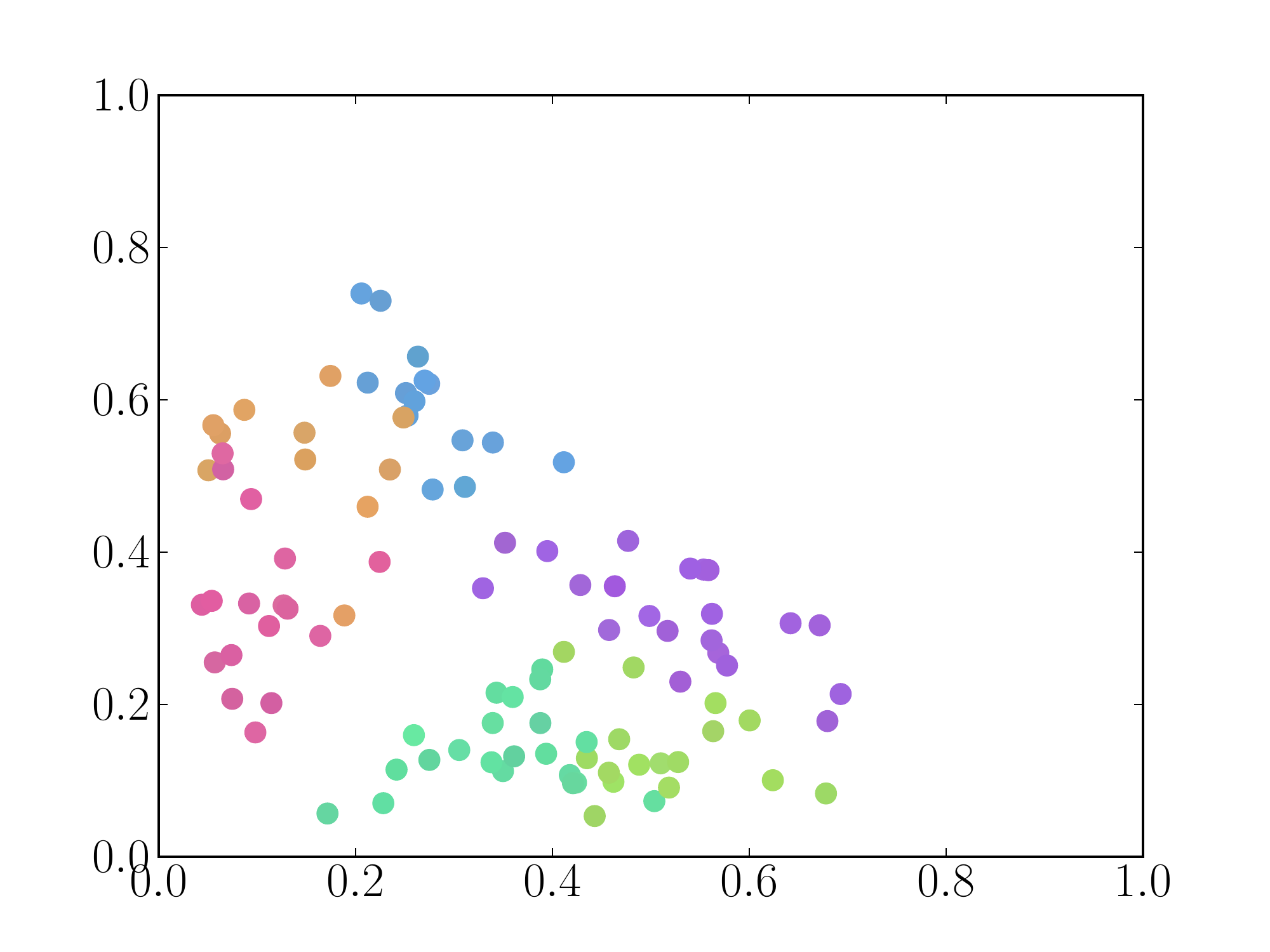}
  \end{subfigure}
  \begin{subfigure}[t]{0.19\textwidth}
    \includegraphics[width=\columnwidth]{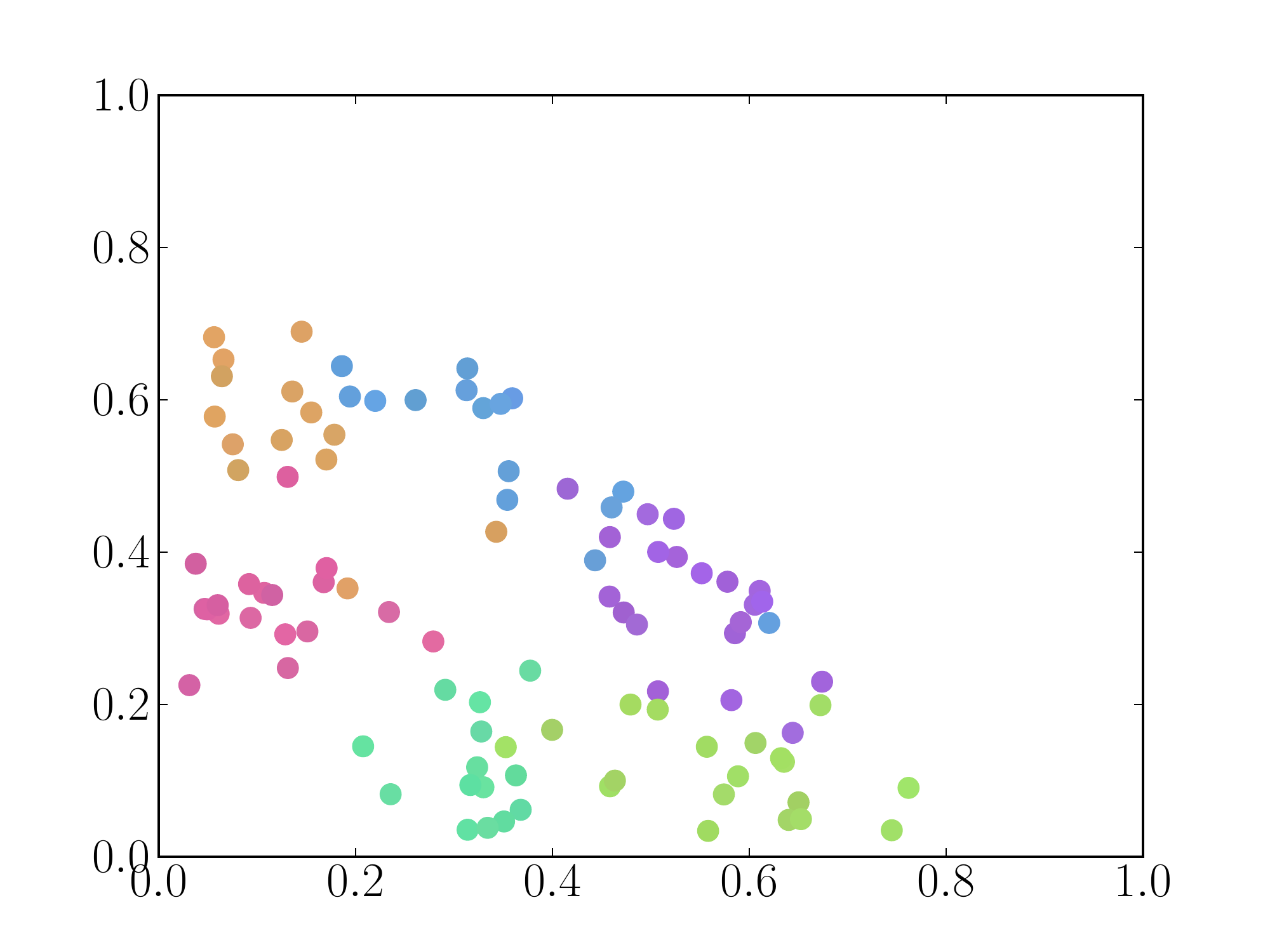}
  \end{subfigure}
  \begin{subfigure}[t]{0.19\textwidth}
    \includegraphics[width=\columnwidth]{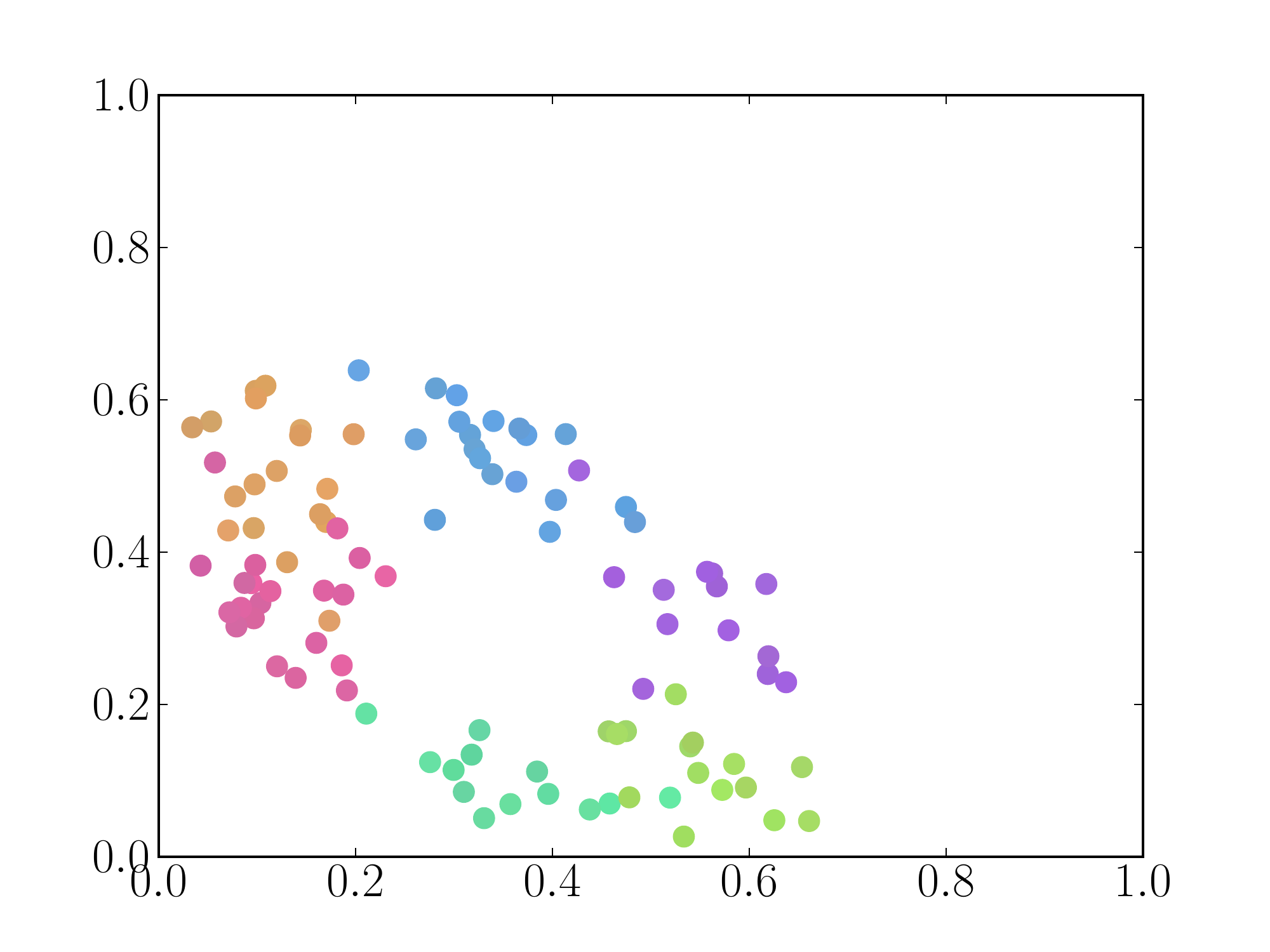}
  \end{subfigure}
  \caption{Samples from the symmetrized batch variational Bayes
  approximate posterior from 5 random restarts. Comparison to Figure
  \ref{fig:postmixtrue} shows that symmetrization reintroduces the
  structure of the true posterior to the approximate posteriors.
}\label{fig:symmetrizedpostmixvb}
\end{figure*}

In order to properly combine the approximate posterior produced by each learning agent,
first the individual posteriors are \emph{symmetrized} (represented by a tilde) by summing over all possible
permutations as follows:
\begin{align}
  \tilde{q}_i(\cdot) &\propto
  \sum_{P}\prod_j q_{P\lambda_{ij}}(\theta_j),
\end{align}
where the sum is taken to be over all permutation matrices $P$
with the same dimension as $\lambda_{ij}$. 
This process of approximating the true single-agent posterior is referred to as symmetrization because $\tilde{q}_i$ has the same 
parameter permutation symmetry as the true posterior, i.e.~
for all permutation matrices $P$,
\begin{align}
  \tilde{q}_i(P\theta_1, \dots,  \dots, P\theta_K) &=
  \tilde{q}_i(\dots).
\end{align}
To demonstrate the effect of this procedure, the mixture model
example was rerun with batch variational Bayesian inference (i.e.~all 30
datapoints were given to a single learner) followed by
symmetrization.
Samples generated from these new symmetrized posterior distributions
over 5 random restarts of the inference procedure 
are shown in Figure \ref{fig:symmetrizedpostmixvb}.
This result demonstrates that the symmetrized distributions are 
invariant to the random permutation to which
the original approximate posterior converged.

It is now possible to combine the individual (symmetrized) posteriors
via the procedure outlined in (\ref{eq:decbayes}):
\begin{align}
  q(\cdot) &\propto q_0(\cdot)^{1-N}\prod_i \tilde{q}_i(\cdot)\nonumber\\
  &= \left(\prod_j
  q_{\lambda_{0j}}(\theta_j)\right)^{1-N}\prod_i\sum_{P_i}\prod_j q_{P_i\lambda_{ij}}(\theta_j)\label{eq:symmdecbayes}\\
  &= \sum_{\{P_i\}_i}\prod_j
  \left[q_{\lambda_{0j}}(\theta_j)^{1-N}\prod_iq_{P_i\lambda_{ij}}(\theta_j)\right],
  \nonumber
\end{align}
where the outer sum is now over unique combinations of the 
set of permutation matrices $\{P_i\}_i$ used by the learning agents.

\subsection{AMPS - APPROXIMATE MERGING OF POSTERIORS WITH SYMMETRY}
The posterior distribution in (\ref{eq:symmdecbayes}) is unfortunately
intractable to use for most purposes, as it contains
a number of terms that is factorial in the dimensions of the parameters, and
exponential in the number of learning agents.
Therefore, we approximate this distribution by finding the component with the
highest weight -- the intuitive reasoning for this is that the component with
the highest weight is the one for which the individual posteriors have correctly
aligned permutations, thus contributing to
each other the most and representing the overall posterior the best. 
While the resulting distribution will not be symmetric,
it will appear as though it were generated from variational Bayesian inference;
this, as mentioned before, is most often fine in practice.

In order to compute the weight of each component, we need to compute
its integral over the parameter space. Suppose that
each approximate posterior component $q_{\lambda_{ij}}(\theta_j)$ has the following form:
\begin{align}
  q_{\lambda_{ij}}(\theta_j) &= h_j(\theta_j) e^{\mathrm{tr}\left[\lambda_{ij}^T
  T(\theta_j)\right] -A_j(\lambda_{ij})}
\end{align}
where $h_j(\cdot)$ and $A_j(\cdot)$ are the base measure and log-partition functions for parameter $j$,
respectively. The trace is used in the exponent in case $\lambda_{ij}$ is
specified as a matrix rather than as a single column vector (such as in the
example presented in Section \ref{subsec:mixegrevisit}). Thus, given a set
of permutation matrices $\{P_i\}_i$, 
the factor of the weight for the component due to parameter $j$ is
\begin{align}
  \begin{aligned}
    W_j\left(\{P_i\}_i\right) &= \int_{\theta_j} q_{\lambda_{0j}}(\theta_j)^{1-N}\prod_i
  q_{P_{i}\lambda_{ij}}(\theta_j).
\end{aligned}\label{eq:compweight}
\end{align}
The overall weight of the component is the product over the parameters,
so finding the maximum weight component of (\ref{eq:symmdecbayes})
is equivalent to finding the set of permutation matrices 
$P^*_i$ that maximizes the product of the $W_j$,
\begin{align}
  \begin{aligned}
    \{P^*_i\}_i \gets \argmax_{\{P_i\}_i} \prod_j W_j\left(\{P_i\}_i\right).
  \end{aligned}\label{eq:basicopt}
\end{align}
Due to the use of exponential family distributions in the
variational approximation,
the optimization (\ref{eq:basicopt}) can be posed as a combinatorial
optimization over permutation matrices with a closed-form objective: 
\begin{align}
  \begin{aligned}
    \max_{\{P_i\}_i}& \sum_j A_j\left( (1-N)\lambda_{0j}
    + \sum_{i} P_i \lambda_{ij}\right)\\
    \mathrm{s.t.}\,& P_i \in S \quad \forall i 
  \end{aligned}\label{eq:combopt}
\end{align}
where $S$ is the symmetric group of order equal to the row dimension of the
matrices $\lambda_{ij}$. Using the convexity of the log-partition
function $A_j\left(\cdot\right)$, the fact that the objective is affine in
its arguments, and the fact that the vertices of the 
Birkhoff polytope are permutation matrices, one can reformulate
(\ref{eq:basicopt}) as a convex maximization over a polytope:
\begin{align}
  \begin{aligned}
    \max_{\{P_i\}_i}& \sum_j A_j\left( (1-N)\lambda_{0j}
    + \sum_{i} P_i \lambda_{ij}\right)\\
    \mathrm{s.t.}\,& P_i^T\bm{1} = \bm{1}, \quad P_i\bm{1} = \bm{1}, \quad P_i \geq 0
    \quad \forall i
  \end{aligned}\label{eq:convexopt}
\end{align}
where $\bm{1}$ is a vector with all entries equal to $1$.
 Global optimization routines for this problem
are intractable for the problem sizes presented by typical Bayesian
models~\citep{Benson85_NRLQ,Falk86_OR}. Thus, the optimization
must be solved approximately, where the choice of the approximate method 
is dependent on the particular form of $A_j(\cdot)$. 

As mentioned earlier, this optimization was formulated assuming that all the
$\theta_j$ were part of a single coupled permutation symmetry set. However,
if there are multiple subsets of the parameters $\theta_1, \dots \, \theta_K$ 
that have coupled permutation symmetry, an optimization of the form (\ref{eq:convexopt}) 
can be solved for each subset independently. In addition, for any parameter
that does not exhibit permutation symmetry, the original na\"{i}ve 
posterior merging procedure in (\ref{eq:decbayes}) may be used.
These two statements follow from the exponential family mean field assumption
used to construct the individual approximate posteriors $q_i$.

\section{EXPERIMENTS}\label{sec:expts}
All experiments were performed on a computer with an Intel Core i7 processor and
12GB of memory.  
\subsection{DECENTRALIZED MIXTURE MODEL EXAMPLE
REVISITED}\label{subsec:mixegrevisit}
The AMPS decentralized inference scheme was applied to the Gaussian mixture
model example from earlier, with three components having unknown means $\mu = (1.0, -1.0, 3.0)$
and cluster weights $\pi = (0.6, 0.3, 0.1)$, and known variance $\sigma^2 =
0.09$. The prior on each mean $\mu_i$ was Gaussian, with mean $\mu_0 = 0.0$ and
variance $\sigma^2_0 = 2.0$, while the prior on the weights $\pi$ 
was Dirichlet, with parameters $(1.0, 1.0, 1.0)$. 
The dataset consisting of the same 30 datapoints from the earlier trial
was used, where each of 10 learning agents received 3 of the datapoints.
Each learning agent used variational Bayesian inference to find their individual
posteriors $q_i(\mu, \pi)$, and then used AMPS to merge them. The only
communication required between the agents was a single broadcast
of each agent's individual posterior parameters.

In this example, the AMPS objective\footnote{The AMPS objective for each experiment was constructed using
  the log-partition function $A_j(\cdot)$ of the relevant
  exponential family models, which may be found in~\citep{Nielsen11_AX}.} was as follows:
\begin{align}
  \begin{aligned}
    &J_{\mathrm{AMPS}} = - \log\Gamma\left(\sum_{j=1}^3\left(\beta_j+1\right)\right) +\\
  &\phantom{=} \sum_{j=1}^3 -\frac{\eta_j^2}{4\nu_j} -
  \frac{1}{2}\log\left(-2\nu_j\right) +\log\Gamma\left(\beta_j + 1\right)
  \end{aligned}
\end{align}
with
\begin{align}
  \lambda_{i \mu} &=\left[\begin{array}{ccc} \eta_{i1} & \eta_{i2} & \eta_{i3}\\
    \nu_{i1} & \nu_{i2} & \nu_{i3} \end{array}\right]^T, \,
                          i=1, \dots, 10\nonumber\\
 \lambda_{i\pi} &= \left[\begin{array}{ccc} \beta_{i1} & \beta_{i2} &
    \beta_{i3}\end{array}\right]^T, \, i=1, \dots, 10 \nonumber\\
  \lambda_{\mu} &= -9\lambda_{0\mu} + \sum_i P_i \lambda_{i\mu}
   \equiv \left[\begin{array}{ccc} \eta_1 & \eta_2 & \eta_3\\
     \nu_1 & \nu_2 & \nu_3\end{array}\right]^T\nonumber\\
  \lambda_{\pi} &= -9\lambda_{0\pi} + \sum_i P_i \lambda_{i\pi} \equiv \left[\begin{array}{ccc} \beta_1 & \beta_2 & \beta_3\end{array}\right]^T\\
  \lambda_{0\mu} &= \left[\begin{array}{ccc} 0 & 0 & 0\\
                            -0.25 & -0.25 & -0.25 \end{array}\right]^T\nonumber\\
  \lambda_{0\pi} &= \left[\begin{array}{ccc} 0 & 0 &
    0\end{array}\right]^T\nonumber\\
  \beta_{ij} &= \alpha_{ij}-1, \, \, \eta_{ij} = \frac{\mu_{ij}}{\sigma^2_{ij}}, \,
  \,\nu_{ij} = -\frac{1}{2\sigma^2_{ij}}\nonumber
\end{align}
where $\alpha_{ij}$ was agent $i$'s posterior Dirichlet variational parameter 
for cluster $j$, and $\mu_{ij}/\sigma^2_{ij}$ were agent $i$'s posterior normal
variational parameter for cluster $j$. The objective was optimized 
approximately over the $3\times 3$ permutation matrices $P_i$ by proposing swaps
of two rows in $P_i$, accepting swaps that increased $J_{\mathrm{AMPS}}$,
and terminating when no possible swaps increased $J_{\mathrm{AMPS}}$.

\begin{figure*}[t!]
  \captionsetup{font=scriptsize}
  \centering
\begin{subfigure}[t]{0.3\textwidth}
    \includegraphics[width=\columnwidth]{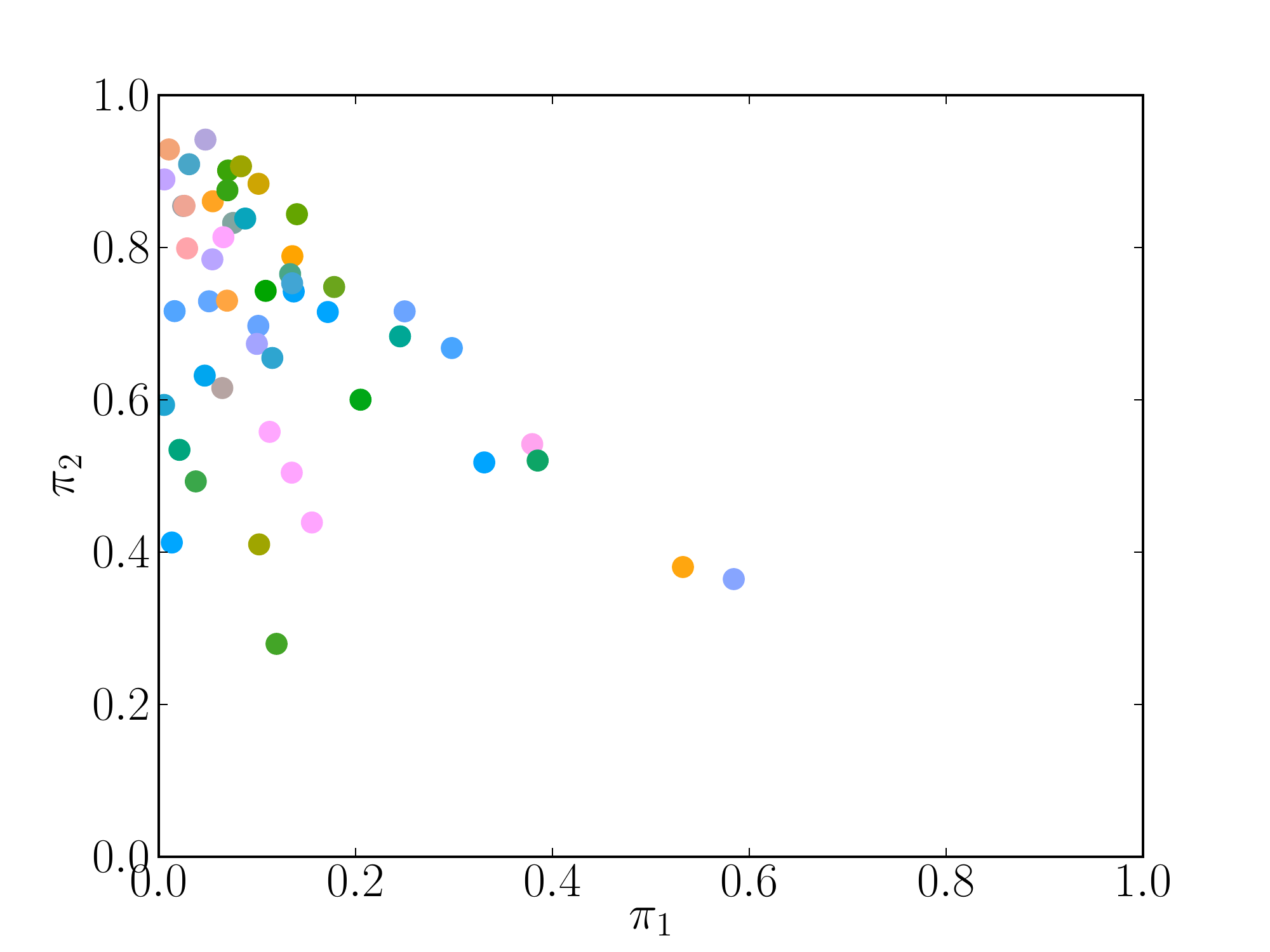}
  \end{subfigure}
  \begin{subfigure}[t]{0.3\textwidth}
    \includegraphics[width=\columnwidth]{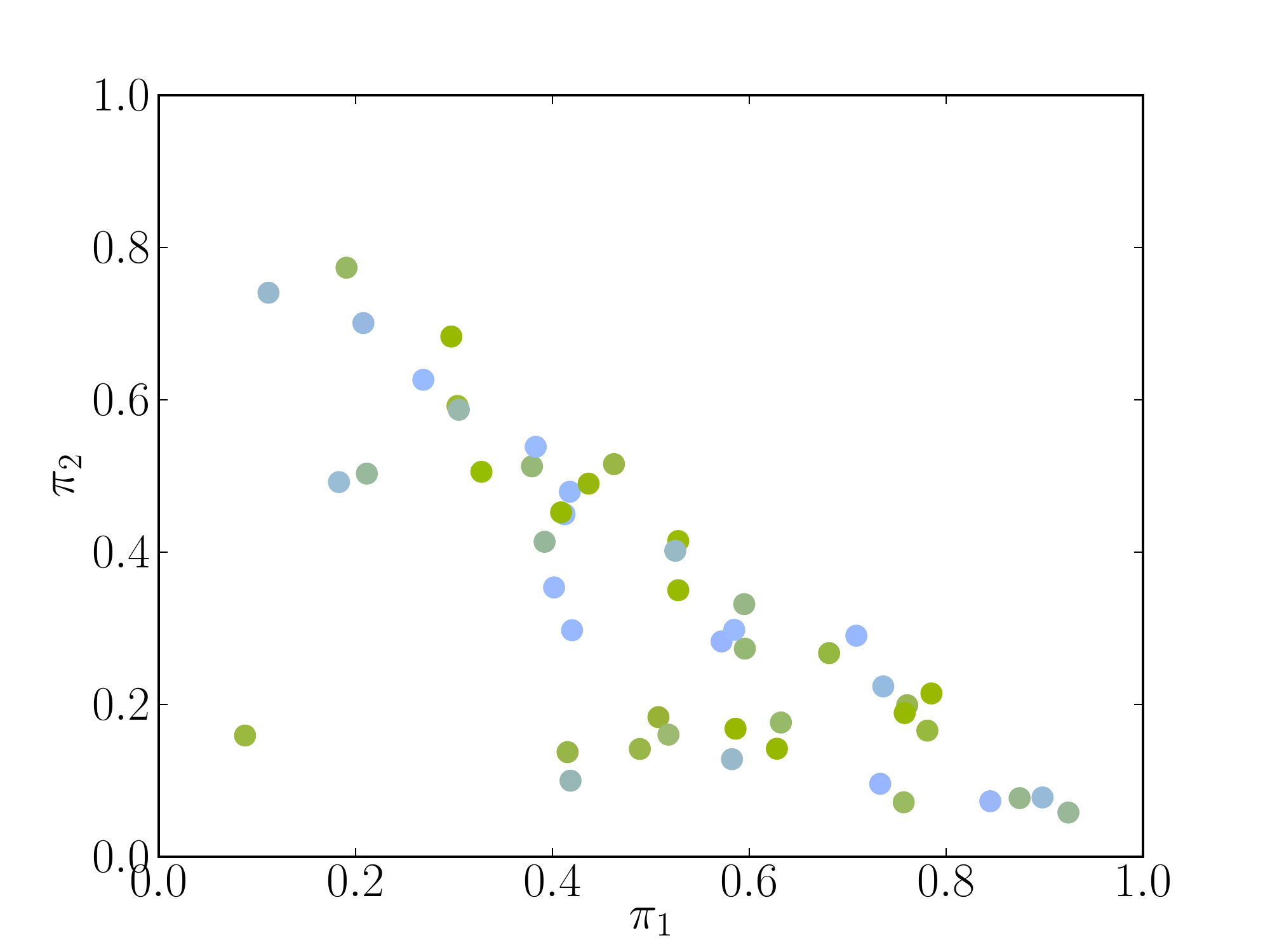}
  \end{subfigure}
  \begin{subfigure}[t]{0.3\textwidth}
    \includegraphics[width=\columnwidth]{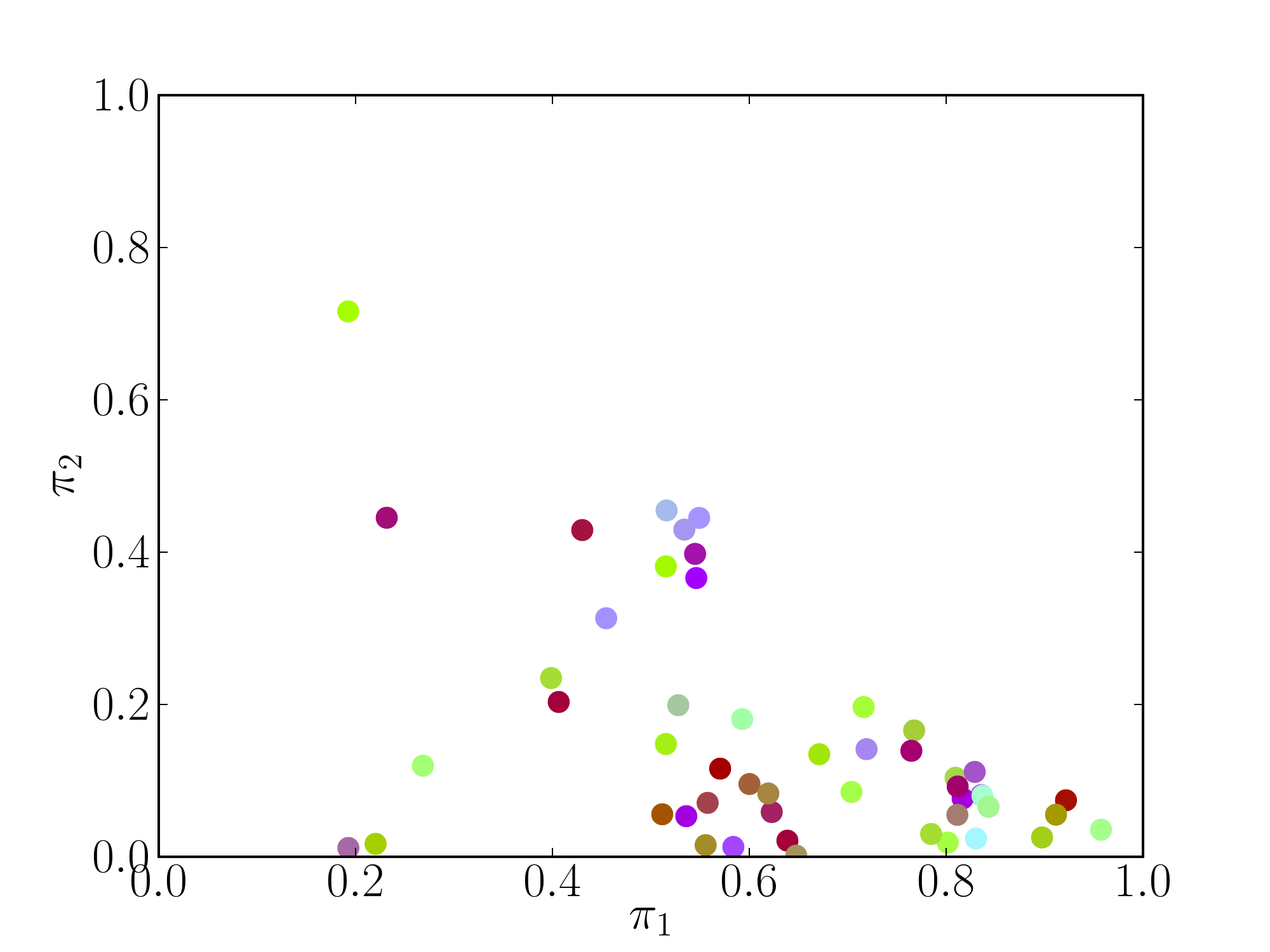}
  \end{subfigure}
  \caption{Samples from the individual posterior distributions from variational Bayesian
    inference for 3 of the learning agents. Note the high level of uncertainty
    in both the weights (position) and cluster locations (colour) in each posterior.
}\label{fig:postmixdecagents}
\end{figure*}

\begin{figure*}[t!]
  \captionsetup{font=scriptsize}
  \centering
  \begin{subfigure}[b]{0.4\textwidth}
    \includegraphics[width=\columnwidth]{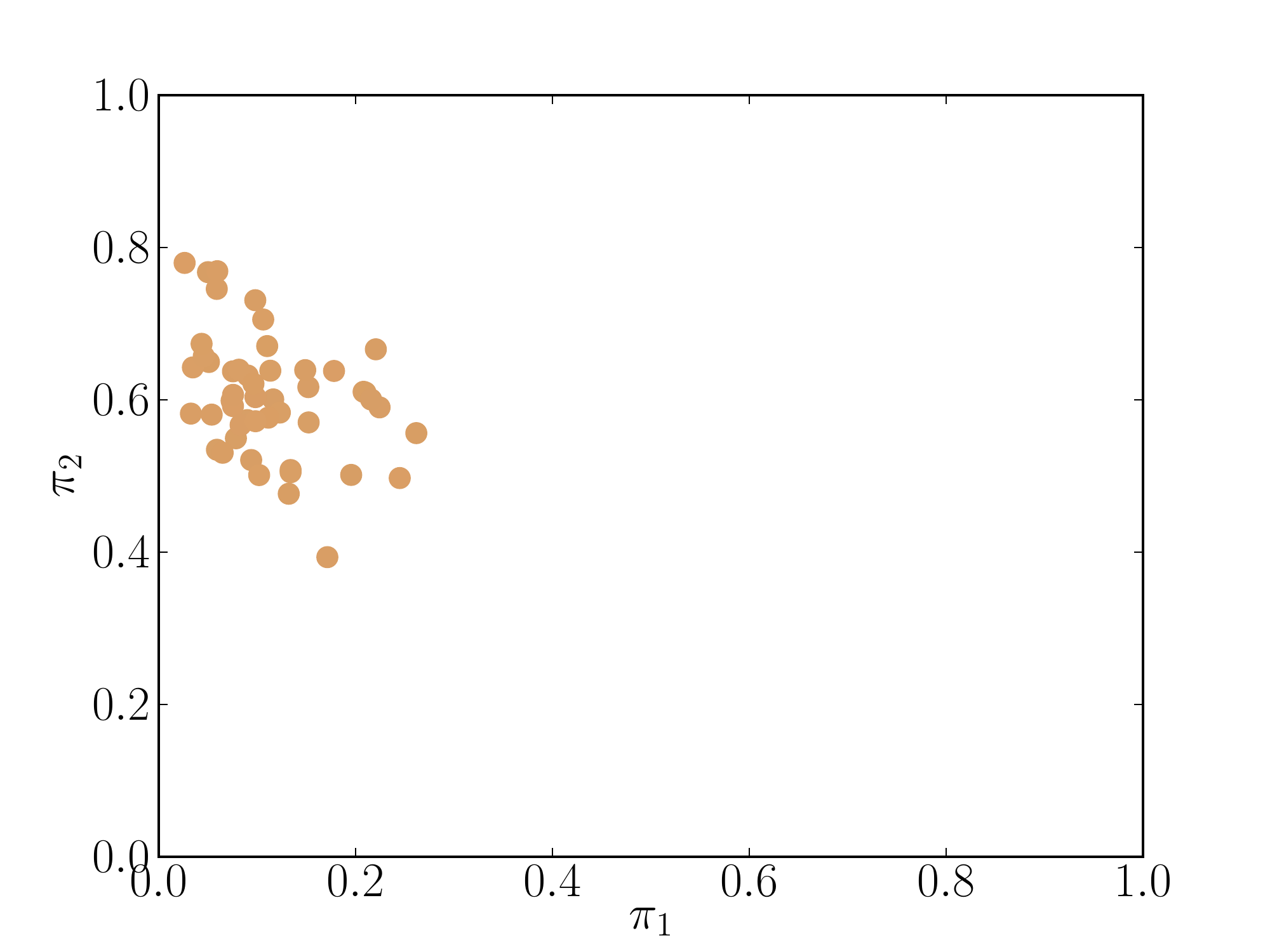}
    \caption{}\label{fig:postmixdecgoodsingle}
  \end{subfigure}
  \begin{subfigure}[b]{0.4\textwidth}
    \includegraphics[width=\columnwidth]{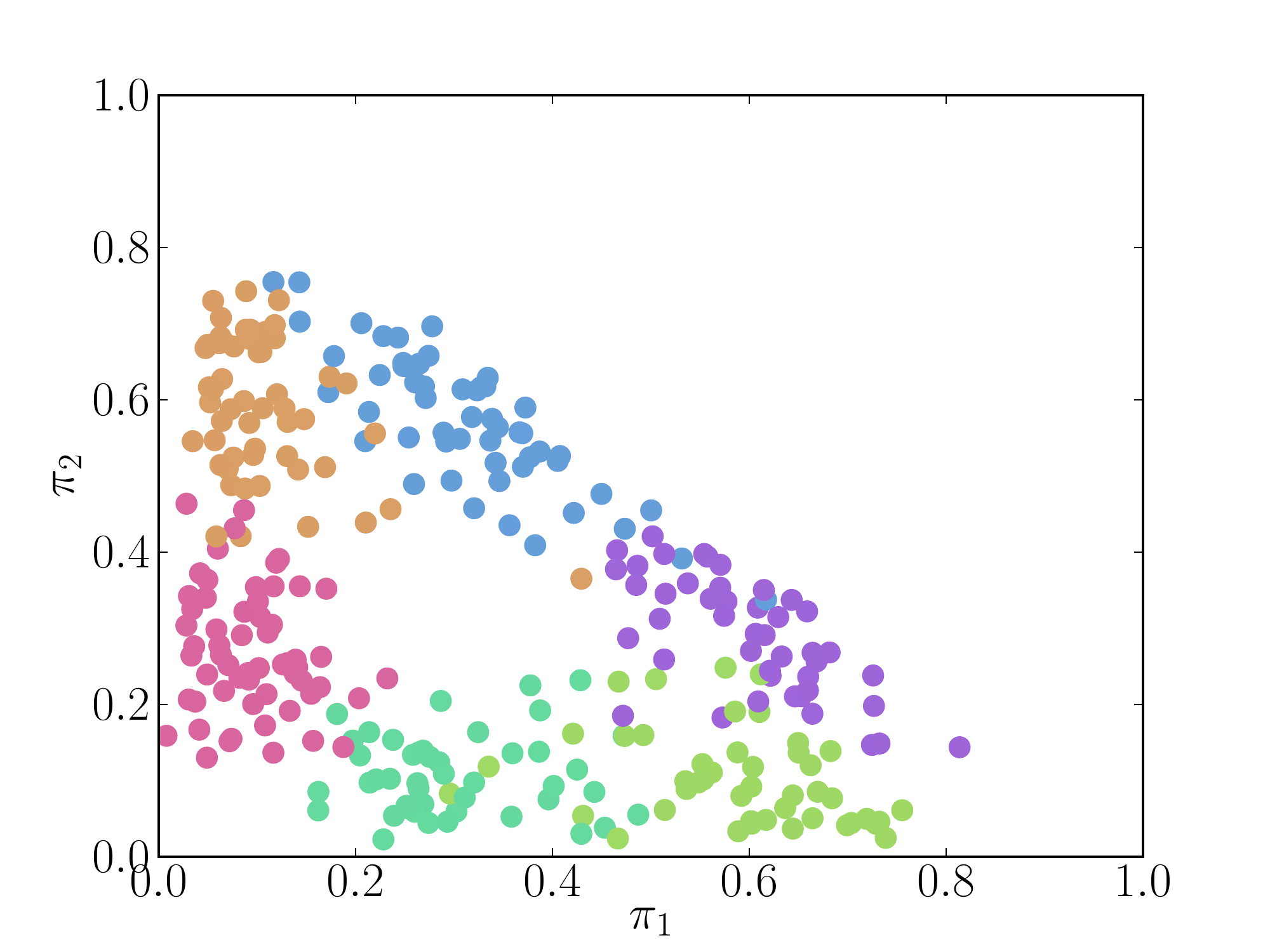}
    \caption{}\label{fig:postmixdecgoodsym}
  \end{subfigure}
  \caption{(\ref{fig:postmixdecgoodsingle}): Samples from the decentralized
  posterior output by AMPS. Comparison to Figure \ref{fig:postmixdecagents}
  shows that the AMPS posterior merging procedure improves the posterior
  possessed by each agent significantly.
  (\ref{fig:postmixdecgoodsym}): Samples from the symmetrized decentralized
posterior. This final symmetrization step is not performed in practice; it is
simply done here for comparison with Figure \ref{fig:postmixtrue}.}\label{fig:postmixdecgood}
\end{figure*}

The individual posteriors for 3 of the learning agents are shown in Figure
\ref{fig:postmixdecagents}, while the decentralized posterior over all the
agents is shown in Figure \ref{fig:postmixdecgood} alongside its symmetrization
(for comparison to the true posterior -- this final symmetrization is not
required in practice). The AMPS posterior is a much better
approximation than the na\"{i}ve decentralized posterior 
shown in Figure \ref{fig:postmixdecbad}; this is because the AMPS posterior
accounts for parameter permutation symmetry in the model prior to
combining the individual posteriors. It may be noted that the decentralized
posterior has slightly more uncertainty in it than the batch posterior,
but this is to be expected when each learning agent individually receives little information
(as demonstrated by the uncertainty in the individual posteriors shown 
in Figure \ref{fig:postmixdecagents}).

\subsection{DECENTRALIZED LATENT DIRICHLET ALLOCATION}\label{subsec:lda}
The next experiment involved running decentralized variational inference
with AMPS on the LDA document clustering model~\citep{Blei03_JMLR}. The dataset
in consideration was the 20 newsgroups dataset, consisting of 18,689 documents with 1,000 held out for
testing, and a vocabulary of 11,175 words after removing stop words and
stemming the remaining words. Algorithms were evaluated based on their 
approximate predictive likelihood of 10\% of the words in each test document given the
remaining 90\%, as described in earlier literature~\citep{Wang11_AISTATS}.
The variational inference algorithms in this experiment 
were initialized using smoothed statistics
from randomly selected documents.

In LDA, the parameter permutation symmetry lies in the arbitrary ordering 
of the global word distributions for each topic. In particular,
for the 20 newsgroups dataset, decentralized learning agents may learn the 20
Dirichlet distributions with a different ordering; therefore, in order to combine the local 
posteriors, we use AMPS with the following objective to reorder each agent's
global topics:
\begin{align}
  \begin{aligned}
    &J_{\mathrm{AMPS}} = \sum_{k=1}^K J_{\mathrm{AMPS}, k} =\\
  &\sum_{k=1}^{K} \sum_{w=1}^{W}
  \log\Gamma\left(\alpha_{kw}\right) -
  \log\Gamma\left(\sum_{w=1}^{W}\alpha_{kw}\right)\\
\end{aligned}\\
\begin{aligned}
  &\alpha = (1-N)\alpha_0 + \sum_{i=1}^N P_i \alpha_i, \quad \alpha, \alpha_i \in
  \mathbb{R}^{K \times W}\\
  &\alpha_{0kw} = \frac{10}{W}\quad K = 20, \quad W = 11,175
  \end{aligned}\nonumber
\end{align}
where $\alpha_{ikw}$ is agent $i$'s posterior Dirichlet variational
parameter for topic $k$ and word $w$, and the optimization
is over $K\times K$ permutation matrices $P_i, i=1, \dots, N$.
For the LDA model, the AMPS objective $J_{\mathrm{AMPS}}$ is additive over 
the topics $k$; therefore, $J_{\mathrm{AMPS}}$ can be optimized
approximately by iteratively solving 
maximum-weight bipartite matching problems as follows:
\begin{enumerate}
  \item{Initialize the decentralized posterior parameter $\alpha \gets
      (1-N)\alpha_0 + \sum_{i=1}^N P_i \alpha_i$ with a set of
  $P_i$ matrices}
\item{For each agent $i$ until $J_{\mathrm{AMPS}}$ stops increasing:
  \begin{enumerate}
    \item{Deassign agent $i$'s posterior: $\alpha \gets \alpha - P_i \alpha_i$}
    \item{Form a bipartite graph with decentralized topics $k$ on one side,
        agent $i$'s topics $k'$ on the other, and edge weights
        $w_{kk'}$ equal to $J_{\mathrm{AMPS}, k}$ if agent $i$'s topic $k'$ is
      assigned to the decentralized topic $k$}
    \item{$P_i \gets$ Maximum weight assignment of agent $i$'s topics}
    \item{Reassign agent $i$'s posterior: $\alpha \gets \alpha + P_i \alpha_i$}
  \end{enumerate} }
\end{enumerate}

\begin{figure*}[t!]
  \captionsetup{font=scriptsize}
  \begin{subfigure}[b]{0.32\textwidth}
    \includegraphics[width=\columnwidth]{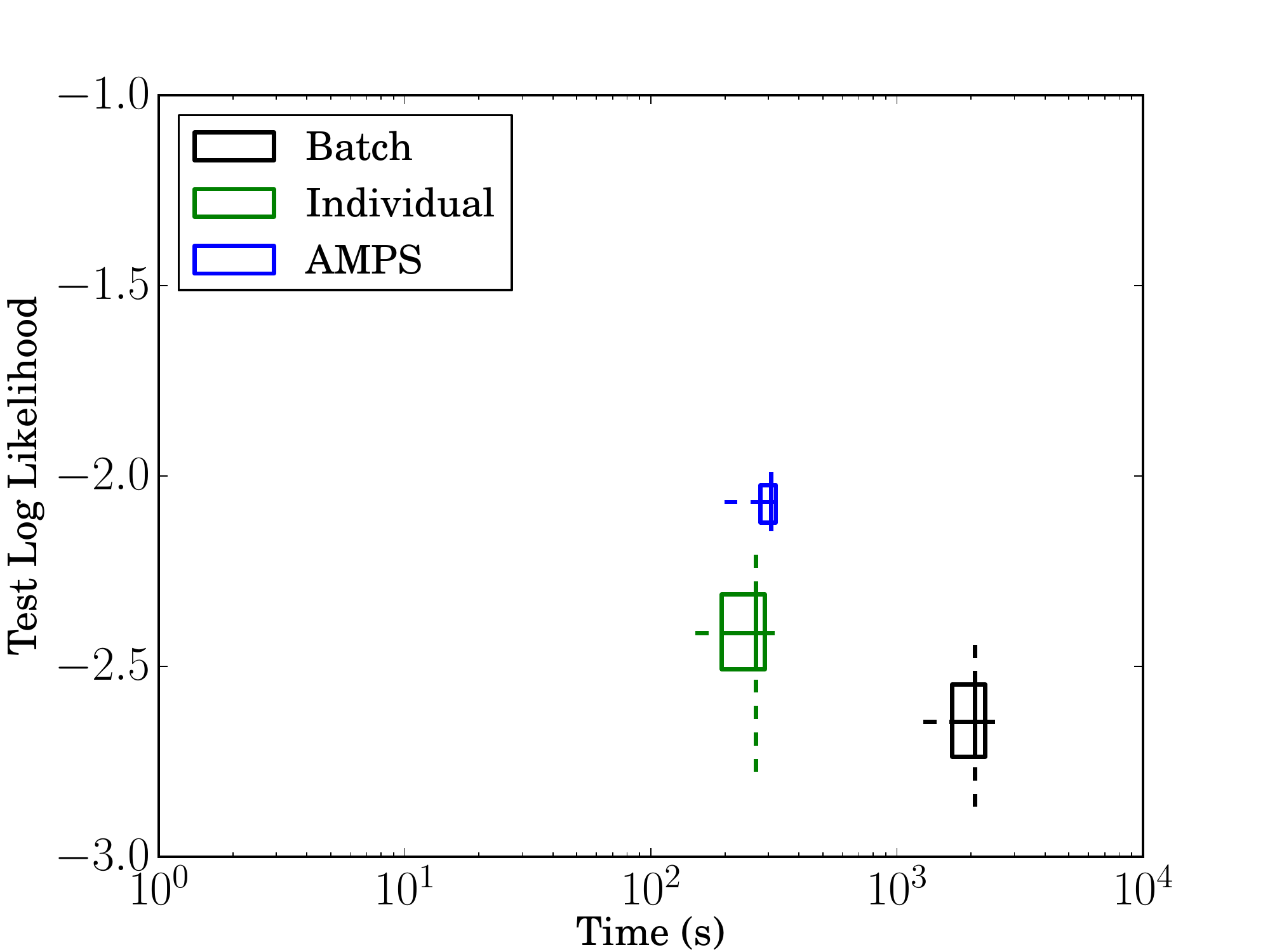}\vspace*{-.05in}
  \captionsetup{font=scriptsize}
    \caption{5 Learning Agents}\label{fig:lda5}
  \end{subfigure}
  \begin{subfigure}[b]{0.32\textwidth}
    \includegraphics[width=\columnwidth]{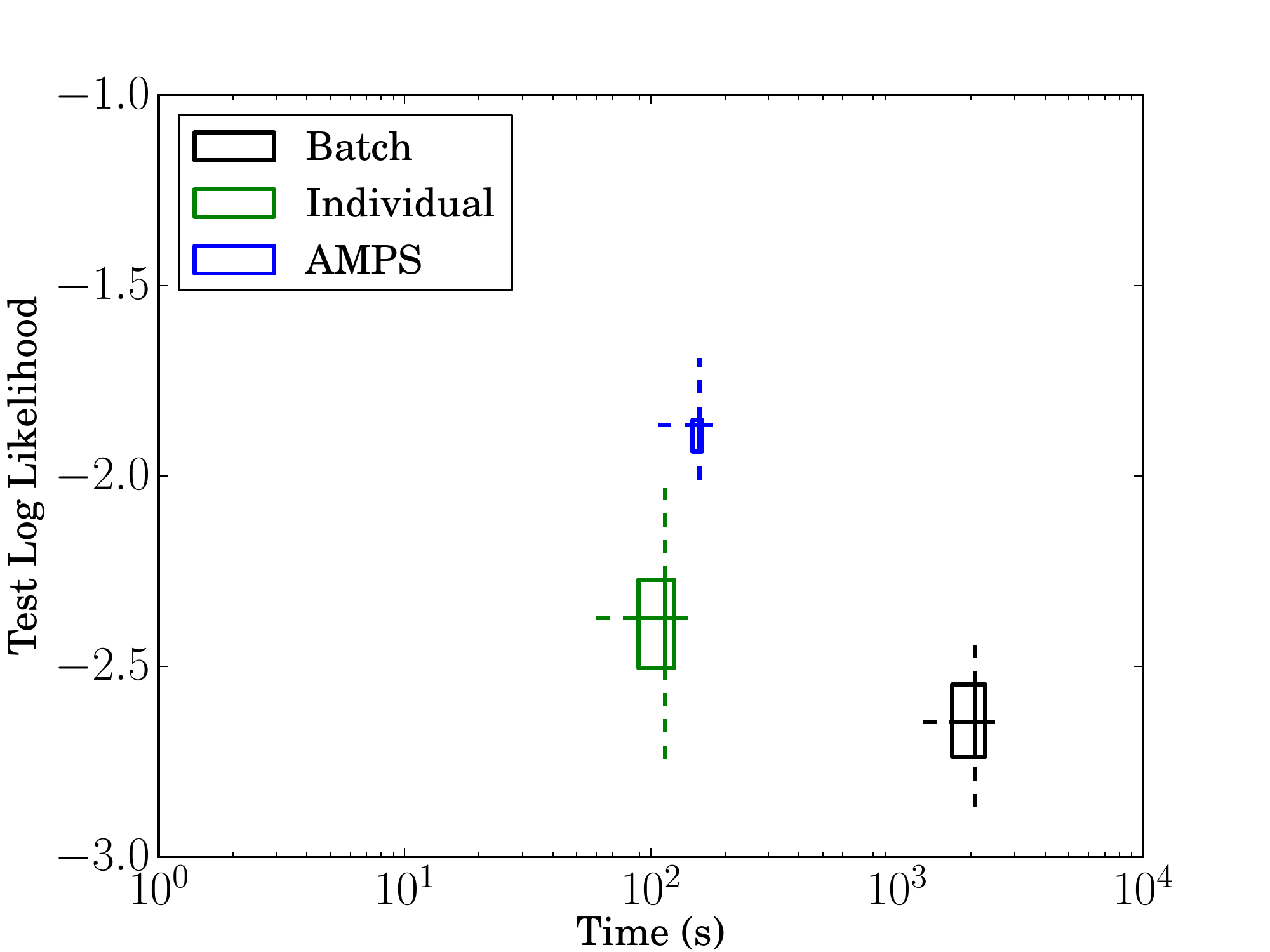}\vspace*{-.05in}
  \captionsetup{font=scriptsize}
    \caption{10 Learning Agents}\label{fig:lda10}
  \end{subfigure}
  \begin{subfigure}[b]{0.32\textwidth}
    \includegraphics[width=\columnwidth]{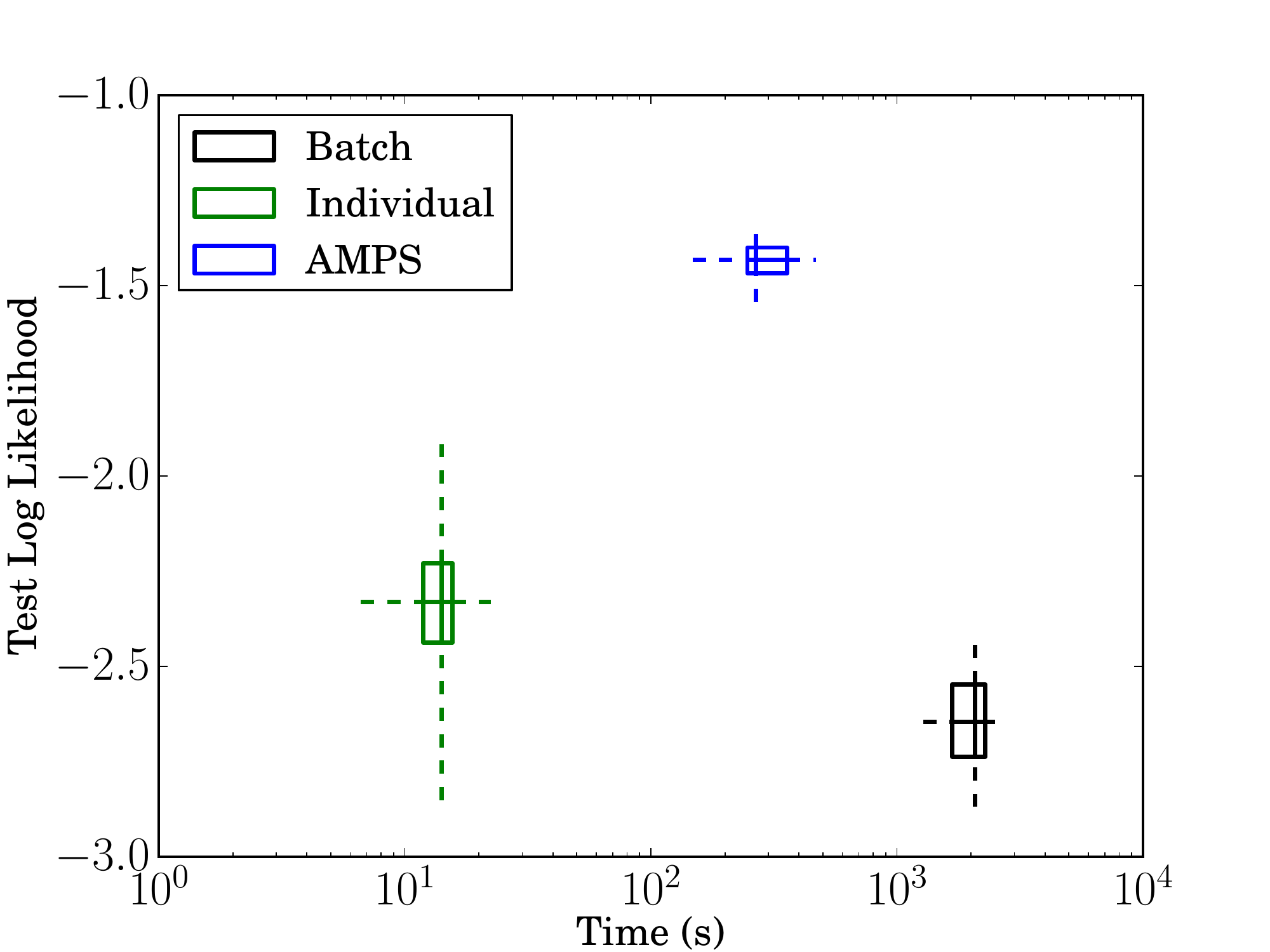}\vspace*{-.05in}
  \captionsetup{font=scriptsize}
    \caption{50 Learning Agents}\label{fig:lda50}
  \end{subfigure}
\vspace*{-.1in}  \caption{Plots of log likelihood on the test data for 5 (\ref{fig:lda5}),
  10 (\ref{fig:lda10}), and  50 (\ref{fig:lda50}) learning agents. The boxes
  bound the $25^\text{th}$ and $75^\text{th}$ percentiles of the data with
  medians shown in the interior, and whiskers show the maximum and minimum values.}\label{fig:ldaresults}
\end{figure*}

First, the performance of decentralized LDA with AMPS was compared to the
batch approximate LDA posterior with a varying number of learning agents.
Figure \ref{fig:ldaresults} shows the test data log likelihood and computation
time over 20 trials for the
batch posterior, the AMPS decentralized posterior, and each individual
agent's posterior for 5, 10, and 50 learning agents.
The results mimic those of the synthetic experiment -- the posterior output by
AMPS significantly outperforms each individual agent's posterior, and the
 effect is magnified as the number of agents increases.
Further, there is a much lower variance in the AMPS posterior test log
likelihood than for each individual agent. The batch method tends to
get stuck in poor local optima in the variational objective, leading to relatively poor performance,
while the decentralized method avoids these pitfalls by solving a number of
smaller optimizations and combining the results afterwards with AMPS. Finally, as the number of agents
increases, the amount of time required to solve the AMPS optimization increases;
reducing this computation time is a potential future goal for research on this
inference scheme.

The next test compared the performance of AMPS to
SDA-Bayes~\citep{Broderick13_NIPS}, a recent streaming, distributed variational 
inference algorithm. The algorithms were tested on  20 trials of each of three settings: one with 1 agent and 10 subbatches of data
per agent; one with 10 agents and 1 subbatch of data per agent; and finally, one with 10 agents and
10 subbatches of data per agent. Each agent processed its subbatches in serial.
For SDA-Bayes, each agent updated a single distributed posterior after each
subbatch. For the decentralized method, each agent used
AMPS to combine the posteriors from its own subbatches, and then used AMPS again
to combine each agent's resulting posterior.

\begin{figure}[t!]
\vspace*{-.1in}  \captionsetup{font=scriptsize}
  \centering
  \includegraphics[width=0.8\columnwidth]{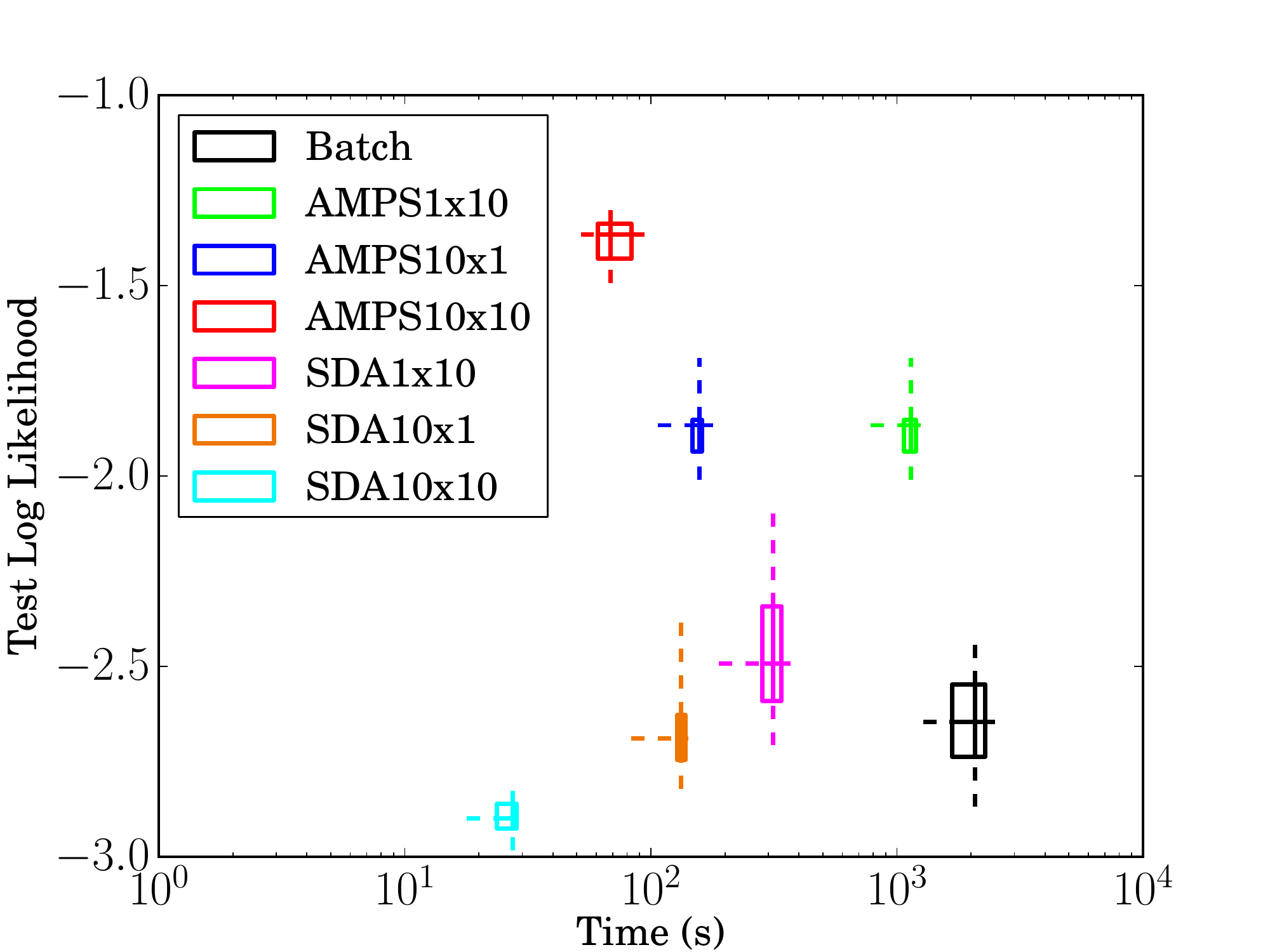}
\vspace*{-.1in}  \caption{Comparison with SDA-Bayes. The $A\times B$ in the
  legend names refer to using $A$ learning agents, where each
  splits their individual batches of
  data into $B$ subbatches.}\label{fig:sdaamps}
\end{figure}

Figure \ref{fig:sdaamps} shows the results from this procedure. AMPS outperforms SDA-Bayes in terms of test log likelihood, and
is competetive in terms of the amount of time it takes to perform inference and
then optimize the AMPS objective. This occurs because AMPS takes into account the arbitrary
ordering of the topics, while SDA-Bayes ignores this when combining posteriors. An interesting note is that the AMPS10x10
result took less time to compute than the time for 50 agents in Figure
\ref{fig:lda50}, despite the fact that it effectively merged 100
posterior distributions; 
this hints that developing a hierarchical optimization scheme for AMPS
is a good avenue for further exploration. A final note is that using AMPS as
described above is not truely a streaming procedure; however, one can rectify
this by periodically merging posteriors using AMPS to form the prior
for inference on subsequent batches.


\subsection{DECENTRALIZED LATENT FEATURE ASSIGNMENT}
The last experiment involved running decentralized variational 
inference with AMPS on a finite latent feature assignment
model~\citep{Griffiths05_NIPS}. In this model, a set of $K$ feature vectors $\mu_k \in
\mathbb{R}^D$ are sampled from a Gaussian prior
$\mu_k \sim \mathcal{N}(0, \sigma_0^2I)$, and a set of feature inclusion probabilities
are sampled from a beta prior $\pi_k \sim \mathrm{Beta}(\alpha_k, \beta_k)$.
Finally, for each image $i$, a set of features $z_i$ are sampled independently from the
weights $z_{ik} \sim \mathrm{Be}(\pi_k)$, and the image $y_i\in\mathbb{R}^D$ is sampled
from a Gaussian likelihood $y_i \sim \mathcal{N}(\sum_{k} \mu_k z_{ik},
\sigma^2I)$. 

Two datasets were used in this experiment. The first was a
synthetic dataset with $K=5$ randomly generated $D=10$-dimensional binary
feature vectors, feature weights sampled uniformly, and 1300 observations sampled with variance $\sigma^2 =
0.04$, with 300 held out for testing.
For this dataset, algorithms were evaluated based on the error between
the means of the feature posteriors and the true set of latent features, and
based on their approximate predictive likelihoods
of a random component in each test observation vector given the other 9
components. The second dataset was a combination of the Yale~\citep{Belhumeur97_IEEEPAMI} and
Caltech\footnote{Available online: http://www.vision.caltech.edu/html-files/archive.html} 
faces datasets, with 581 $32\times32$ frontal images of faces, where 50 were held out for
testing. The number of latent features was set to $K=10$. 
For this dataset, algorithms were evaluated based on their approximate predictive
likelihood of 10\% of the pixels in each test image given the remaining 90\%,
and the inference algorithms were initialized using
smoothed statistics from randomly selected images.
\begin{figure*}[ht!]
  \captionsetup{font=scriptsize}
  \centering
  \begin{subfigure}[b]{0.32\textwidth}
    \includegraphics[width=\columnwidth]{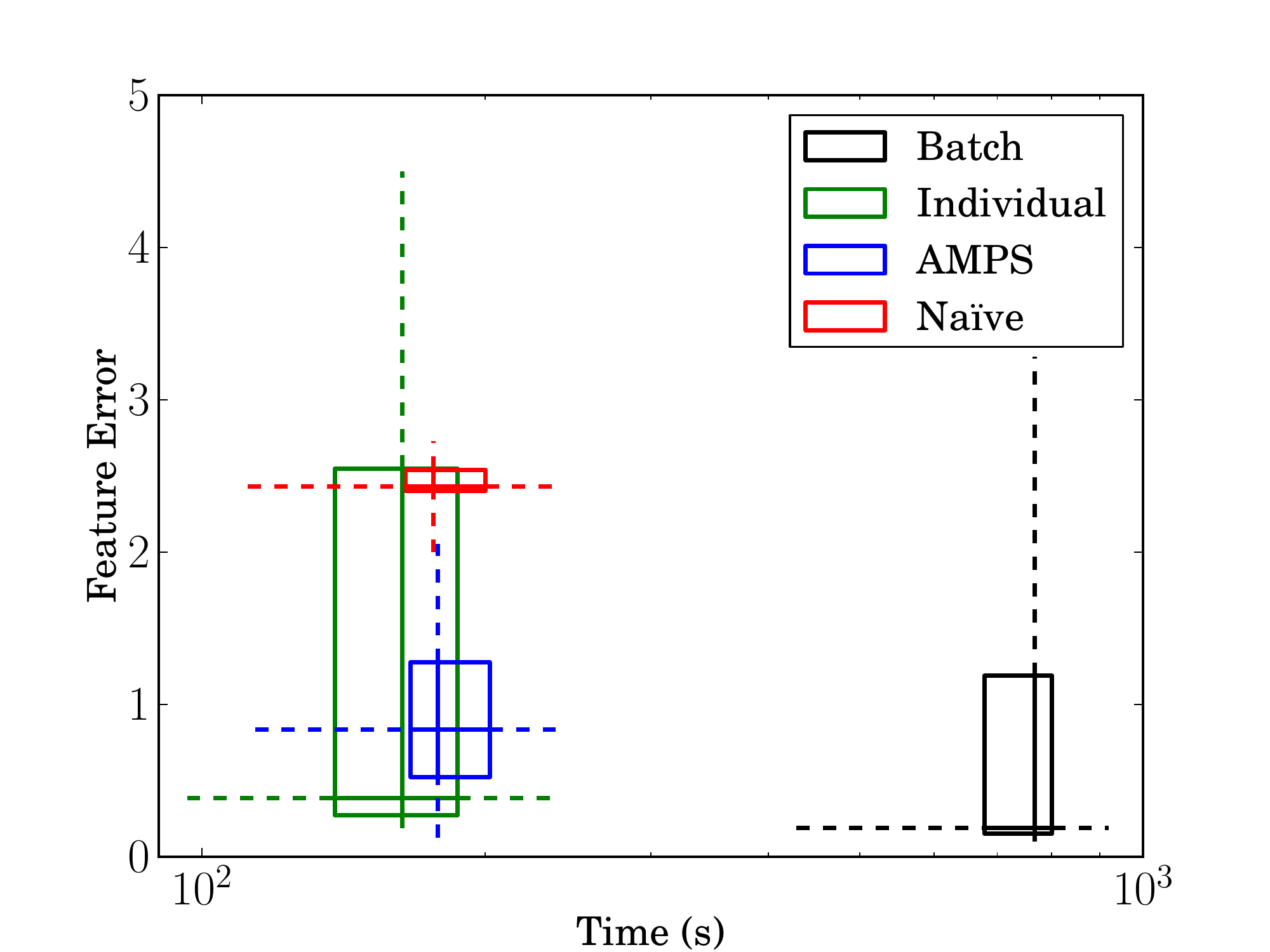}
  \captionsetup{font=scriptsize}
    \caption{}\label{fig:synthlfa}
  \end{subfigure}
  \begin{subfigure}[b]{0.32\textwidth}
    \includegraphics[width=\columnwidth]{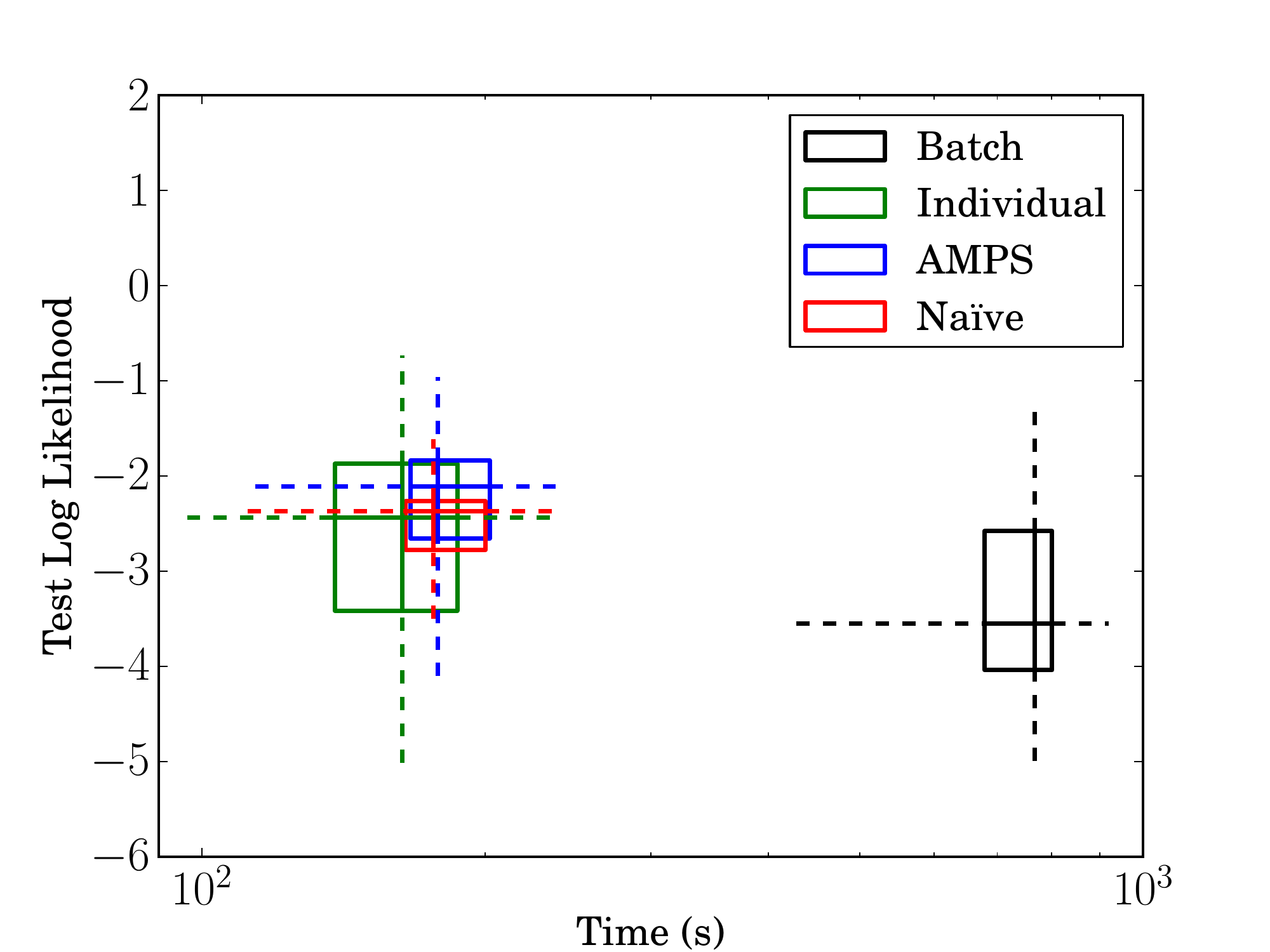}
  \captionsetup{font=scriptsize}
    \caption{}\label{fig:synthlfatll}
  \end{subfigure}
  \begin{subfigure}[b]{0.32\textwidth}
    \includegraphics[width=\columnwidth]{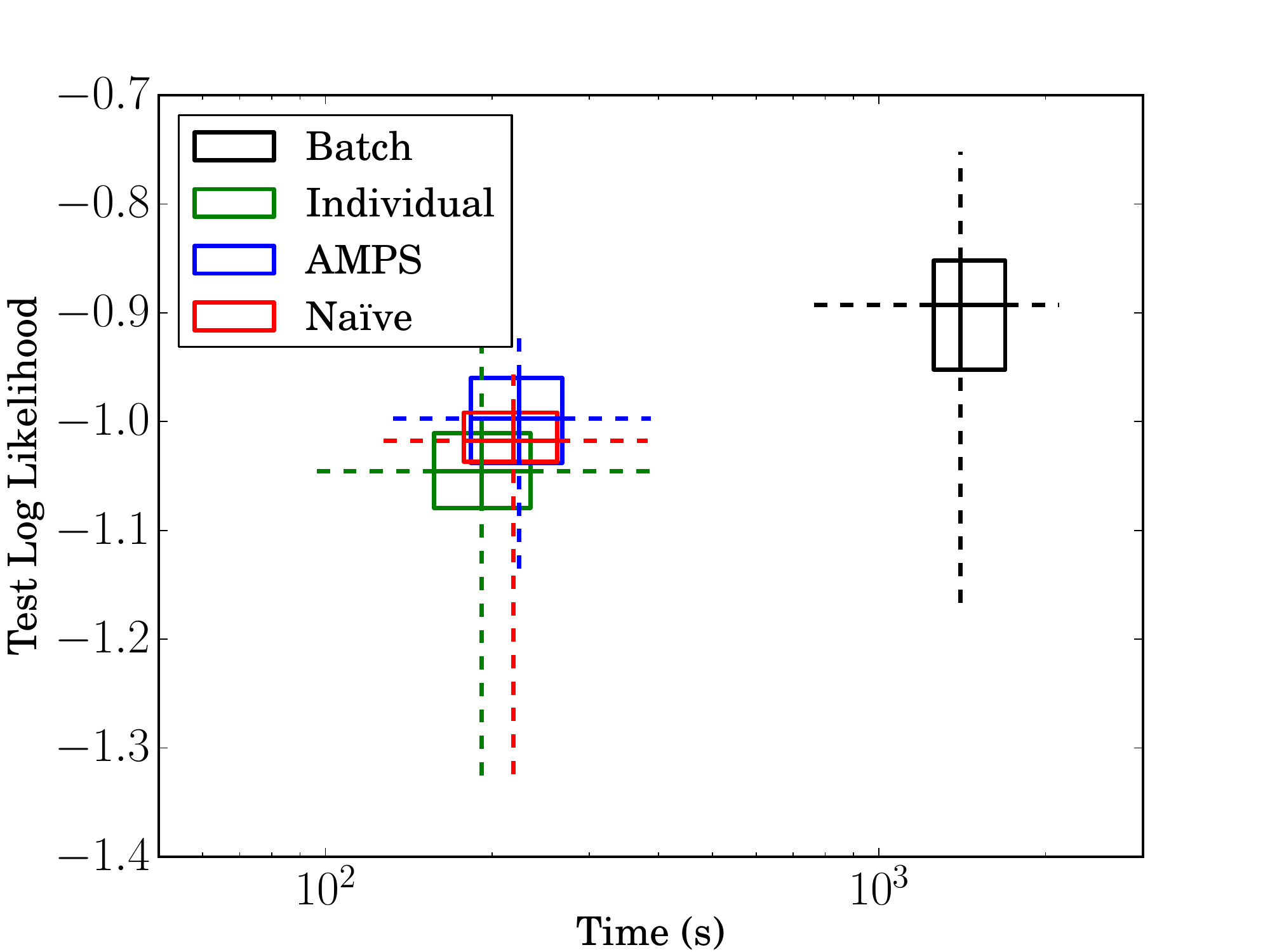}
  \captionsetup{font=scriptsize}
    \caption{}\label{fig:facetll}
  \end{subfigure}
  \caption{(\ref{fig:synthlfa}): 2-norm error between the discovered features and the
  true set for the synthetic dataset. (\ref{fig:synthlfatll}): Test log likelihood 
  for the synthetic dataset. (\ref{fig:facetll}):
Test log likelihood on the faces dataset. All distributed/decentralized results
were generated using 5 learning agents.}\label{fig:lfa}
\end{figure*}

The parameter permutation symmetry in the posterior of the latent feature model lies in the
ordering of the features $\mu_k$ and weights $\pi_k$. Therefore,
to combine the local posteriors, we use AMPS with the following objective to
reorder each agent's set of features and weights:
\begin{align}
  \begin{aligned}
    &J_{\text{AMPS}} = \sum_{k=1}^K J_{\text{AMPS}, k} = \\
    &\sum_{k=1}^K\log\Gamma(\alpha_k) + \log\Gamma(\beta_k) -
    \log\Gamma(\alpha_k+\beta_k)  \\
    &\phantom{\sum_{k=1}^K}-\frac{\eta_k^T\eta_k}{4\nu_k} -
    \frac{D}{2}\log(-2\nu_k)
  \end{aligned}
\end{align}
where $\alpha, \beta \in \mathbb{R}^K$ are the combined posterior beta
natural parameters, and $\eta\in\mathbb{R}^{D\times K}, \nu\in\mathbb{R}^K$ are the
combined posterior normal natural
parameters (combined using the $(1-N)*_0 + \sum_i P_i*_i$ rule as described in
the foregoing). The priors were $\alpha_{k0} = \beta_{k0} = 1$, $\eta_{k0} =
0\in\mathbb{R}^D$, and $\nu_{k0}$ was estimated from the data. As in the LDA model, the AMPS objective for the latent
feature model is additive over the features $k$;
therefore the optimization was performed using iterative maximum-weight bipartite
matchings as described in Section \ref{subsec:lda}. 

Figure \ref{fig:lfa} shows the results from the two datasets
using batch learning and decentralized learning. For the decentralized
results, the posteriors of 5 learning agents were combined using AMPS
or the na\"{i}ve approach (equivalent to SDA5$\times$1 in the notation 
of Figure \ref{fig:sdaamps}). Figure \ref{fig:synthlfa} shows that 
AMPS discovers the true set of latent features with a lower 2-norm error
than both the na\"{i}ve posterior combination and the individual learning
agents, with a comparable error to the batch learning case.
However, as shown in Figures \ref{fig:synthlfatll} (synthetic) and
\ref{fig:facetll} (faces), AMPS only outperforms the na\"{i}ve approach in terms
of predictive log likelihoods on the held-out test set by a small margin. 
This is due to the flexibility of the latent feature assignment model, in
that there are many sets of latent features that explain the observations well.

\section{DISCUSSION}
This work introduced the Approximate Merging of Posteriors with Symmetry (AMPS)
algorithm for approximate decentralized variational inference. AMPS may be used
in ad-hoc, asynchronous, and dynamic networks. Experiments demonstrated the 
modelling and computational advantages of AMPS with respect to batch
and distributed learning. Motivated by 
the examples in Section \ref{sec:expts}, there is certainly room for
improvement of the AMPS algorithm. For example, it may be possible to reduce the
computational cost of AMPS by using a hierarchical optimization scheme,
rather than the monolithic approach used in most of the examples presented in the
foregoing. 
Further, extending AMPS for use with Bayesian nonparametric models
is of interest for cases when the number of latent parameters is unknown a priori, or
when there is the possibility that agents learn disparate sets of
latent parameters that are not well-combined by optimizing over permutations.
Finally, while the approximate optimization algorithms presented herein work well in practice, it would be of interest to find
 bounds on the performance of such algorithms with respect to the
true AMPS optimal solution.

\subsubsection*{Acknowledgements}
This work was supported by the Office of Naval Research under ONR MURI grant N000141110688.
\begingroup
\renewcommand{\section}[2]{\subsubsection*{#2}}
\bibliographystyle{abbrvnat}
{\small
\bibliography{main}
}
\endgroup

\end{document}